\documentclass[conference]{IEEEtran}
\usepackage[colorlinks=true,citecolor=blue,linkcolor=blue,urlcolor=blue,bookmarks=true]{hyperref}
\usepackage[numbers, sort]{natbib}
\usepackage{adjustbox}
\usepackage{algorithm}
\usepackage{amsfonts}
\usepackage{amsmath}
\usepackage{amssymb}
\usepackage{array, booktabs, makecell}
\usepackage{babel,blindtext}
\usepackage{braket}
\usepackage{graphicx}
\usepackage{multicol}
\usepackage{multirow}
\usepackage{siunitx}
\usepackage{subcaption}
\usepackage{tabularx}
\usepackage{times}
\usepackage{varwidth}
\usepackage{xspace}
\usepackage{verbatim}
\usepackage{xcolor}

\makeatletter
\newcommand\fs@spaceruled{\def\@fs@cfont{\bfseries}\let\@fs@capt\floatc@ruled
  \def\@fs@pre{\vspace{0.6\baselineskip}\hrule height.8pt depth0pt \kern2pt}%
  \def\@fs@post{\kern2pt\hrule\relax}%
  \def\@fs@mid{\kern2pt\hrule\kern2pt}%
  \let\@fs@iftopcapt\iftrue}
\makeatother

\newcommand{\algacro}{MINT\xspace}

\newcommand{\vqvae}{SDAT\xspace}
\newcommand{\vla}{MINT-4B\xspace}
\newcommand{\tf}{MINT-30M\xspace}

\newcommand{\secref}[1]{Section~\ref{#1}}

\renewcommand{\eqref}[1]{(\ref{#1})}

\newcommand{\figref}[1]{Fig.~\ref{#1}}

\newcommand{\vqvaefull}{Spectrally Disentangled Action Tokenizer\xspace}
\usepackage{times}

\usepackage[numbers]{natbib}
\usepackage{multicol}
\usepackage[bookmarks=true]{hyperref}
\usepackage{graphicx}
\usepackage{booktabs} 
\usepackage{multirow} 
\usepackage{graphicx} 
\usepackage{array}     
\usepackage[table]{xcolor}
\usepackage{algorithmic}
\usepackage{microtype}
\usepackage{algorithm}
\usepackage{amsmath}	
\usepackage{setspace}
\usepackage{arydshln}
\usepackage{textcomp}
\usepackage{marvosym}
\usepackage{bm}
\usepackage{cuted}
\usepackage{hyperref}
\usepackage{caption}
\usepackage{makecell}
\definecolor{sotagray}{gray}{0.9}      
\definecolor{pi0light}{rgb}{0.95, 0.98, 1.0} 
\definecolor{oursvivid}{rgb}{1.0, 0.9, 0.8}  

\begin{document}

\title{Mimic Intent, Not Just Trajectories}



\title{Mimic Intent, Not Just Trajectories}

\author{
Renming Huang$^{1,2}$,
Chendong Zeng$^{1,2}$,
Wenjing Tang$^{1}$,
Jintian Cai$^{1,2}$,
Cewu Lu$^{1,2}$,
Panpan Cai$^{1,2,\dagger}$\\
$^{1}$Shanghai Jiao Tong University \quad
$^{2}$Shanghai Innovation Institute\\
$^\dagger$Corresponding author \\
\textbf{Project Page: }\textcolor{orange!80}{https://renming-huang.github.io/MINT}
}


%

\maketitle

\begin{strip}
    \vspace*{-1.2cm}
    \centering
    \includegraphics[width=\linewidth]{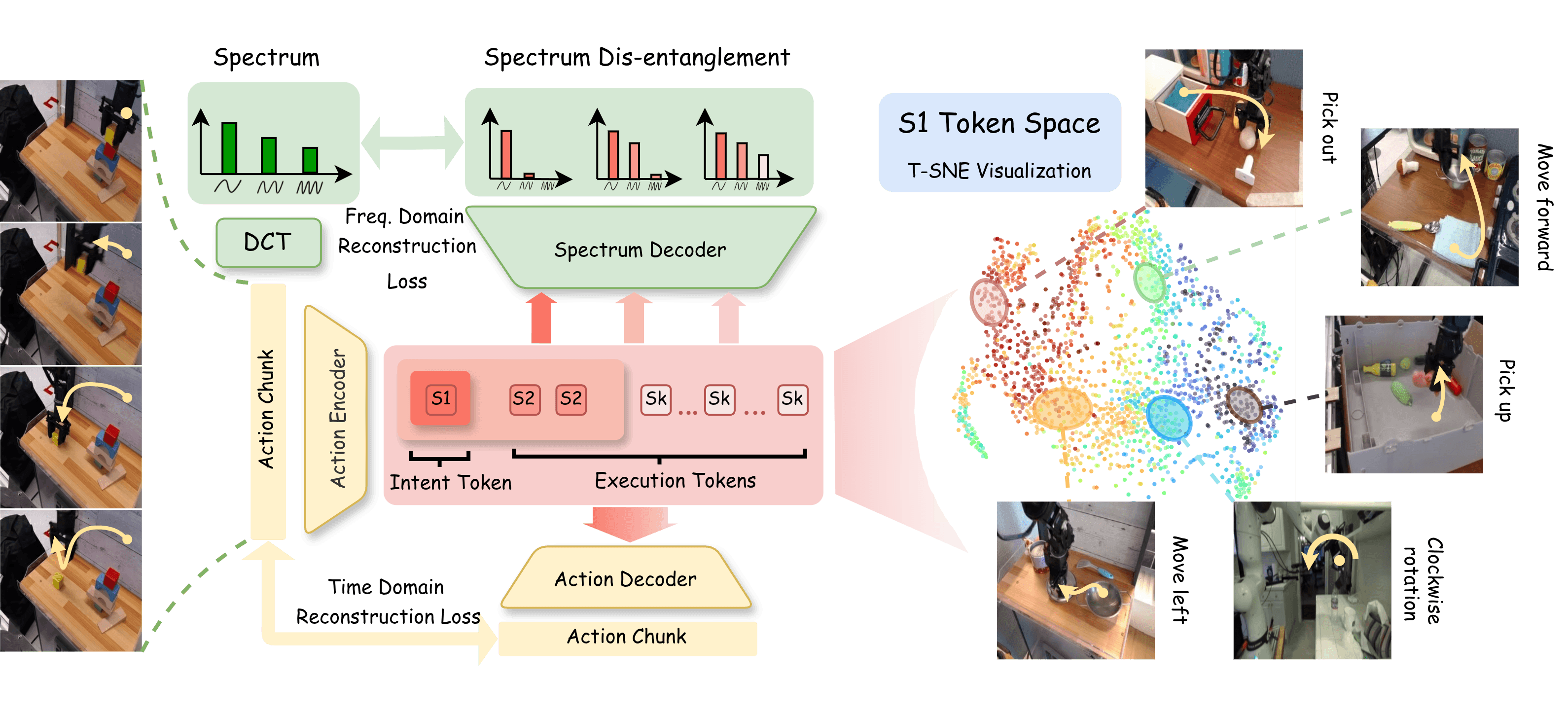}
    \captionof{figure}{\textbf{Left:} We propose Spectrally Disentangled Action Tokenizer, 
    which encodes action chunks into multi-scale tokens via scale-wise frequency domain reconstruction constraints, 
    where the coarsest scale captures global intent and finer scales encode execution residuals. 
    \textbf{Right:} The T-SNE visualization of the $S_1$ token space demonstrates that the learned $S_1$ tokens form distinct clusters corresponding to semantically consistent behaviors (e.g., ``Pick up'',``Move forward'' and ``Clockwise Rotation'')}
    \label{fig:teaser}
\end{strip}

\begin{abstract}
While imitation learning (IL) has achieved impressive success in dexterous manipulation through generative modeling and pretraining, state-of-the-art approaches like Vision-Language-Action (VLA) models still struggle with adaptation to environmental changes and skill transfer. We argue this stems from mimicking raw trajectories without understanding the underlying intent.
To address this, we propose explicitly disentangling behavior intent from execution details in end-2-end IL: \textit{``Mimic Intent, Not just Trajectories'' (MINT)}.
We achieve this via \textit{multi-scale frequency-space tokenization}, which enforces a spectral decomposition of action chunk representation. We learn action tokens with a multi-scale coarse-to-fine structure, and force the coarsest token to capture low-frequency global structure and finer tokens to encode high-frequency details. This yields an abstract \textit{Intent token} that facilitates planning and transfer, and multi-scale \textit{Execution tokens} that enable precise adaptation to environmental dynamics. Building on this hierarchy, our policy generates trajectories through \textit{next-scale autoregression}, performing progressive \textit{intent-to-execution reasoning}, thus boosting learning efficiency and generalization. Crucially, this disentanglement enables \textit{one-shot transfer} of skills, by simply injecting the Intent token from a demonstration into the autoregressive generation process. Experiments on several manipulation benchmarks and on a real robot demonstrate state-of-the-art success rates, superior inference efficiency, robust generalization against disturbances, and effective one-shot transfer.

\end{abstract}

\IEEEpeerreviewmaketitle

\section{Introduction}

Imitation learning from demonstrations has become a dominant paradigm for learning robot manipulation policies. Recent advances are largely driven by vision–language–action (VLA) models~\cite{black2024pi_0,bjorck2025gr00t,kim2024openvla}, which map visual observations and language instructions directly to continuous control commands, achieving impressive performance on dexterous tasks such as folding laundry, pouring coffee, and object rearrangement. However, despite their success in closed settings, these models often generalize poorly to environmental variations and new task instances~\cite{fei2025libero}. We argue that a key limitation is that most existing approaches learn to mimic trajectories as raw signals, without modeling why a particular sequence of actions is executed. As a result, learned policies tend to overfit to surface-level correlations in demonstrations, rather than capturing the underlying behavioral intent that governs task execution.


To address this limitation, recent work has explored action tokenization, which maps continuous trajectories into discrete latent representations~\cite{bu2025univla,wang2025vq,ye2024latent} . Discrete tokens align with the intuition that action semantics are structured and compositional, and token-based policies predict abstract action sequences before decoding them into executable trajectories. However, existing tokenization methods largely function as compression mechanisms~\cite{pertsch2025fast,wang2025vq} rather than semantic abstractions. Their learning objectives are typically agnostic to action meaning, providing no explicit constraint that aligns the token space with interpretable behavioral concepts such as intent. Even when multi-scale or hierarchical tokenization is adopted~\cite{gong2025carp}, the semantics of coarse representations remain unconstrained.

To fill the gap, we introduce \algacro — Mimic Intent, Not just Trajectories, an imitation learning framework based on multi-scale frequency-space action tokenization. \algacro explicitly disentangles behavioral intent from execution details through \textit{spectral decomposition}. The key insight is that a trajectory can be viewed as a superposition of signals at different frequencies: low-frequency components characterize the global shape and long-horizon structure of the behavior, while high-frequency components encode fine-grained execution details and reactive adjustments.

Concretely, we transform action chunks from the time domain into the frequency domain using the Discrete Cosine Transform (DCT)~\cite{ahmed2006discrete}. We train a multi-scale variational autoencoder (VAE)~\cite{tian2024visual} with a frequency-domain reconstruction objective, enforcing consistency between the spectral representations of the original and reconstructed trajectories. The latent space is organized into multiple token scales ($S_1, S_2, \dots, S_k$), with progressively increasing capacity. The coarsest scale contains a single token, while finer scales introduce additional tokens to capture residual information. 

To enforce disentanglement, we design a progressive reconstruction scheme: the model is trained to reconstruct the frequency-domain trajectory using (i) $S1$ alone, (ii) $S_1 + S_2$, (iii) $S_1 + S_2 + S_3, ...$, etc..
This structure induces a clear learning behavior---different levels of abstraction naturally attend to different regions of the frequency spectrum: the $S_1$ token is forced to capture the dominant, low-frequency components to minimize reconstruction error, while finer tokens specialize in modeling high-frequency residuals. 
This spectral separation induces a principled disentanglement between intent and execution, rather than relying on heuristic or post-hoc interpretation of latent variables. We therefore interpret $S_1$ as an ``Intent token'', and $S_2\sim S_K$ as ``Execution tokens''.

This representation enables several key benefits. First, progressive prediction of $S_1\sim S_K$ naturally induces an intent-to-execution reasoning process in latent space, improving sample efficiency and stabilizing long-horizon generation. Second, the Intent token provides a {more} compact, reusable task specification {than} language instructions. Given a single demonstration of a novel task, we can extract its Intent token and inject it into the policy’s autoregressive generation process, enabling one-shot skill transfer to new layouts, new tasks, and extended horizons.

We evaluate \algacro on four manipulation benchmarks, LIBERO~\cite{liu2023libero}, MetaWorld~\cite{yu2020meta}, CALVIN~\cite{mees2022calvin}, and the more challenging LIBERO-Plus~\cite{fei2025libero}, as well as on a real robotic system. \algacro achieves state-of-the-art performance on standard benchmarks, outperforming strong pretrained VLA models ($\pi_{0.5}$~\cite{black2025pi_}), action-tokenization-based methods (UniVLA~\cite{bu2025univla}), and classic imitation learning approaches (ACT~\cite{zhao2023learning}, Diffusion Policy~\cite{chi2025diffusion}). When trained on LIBERO and evaluated on LIBERO-Plus under stronger disturbances, \algacro demonstrates substantially improved robustness, achieving $ 15\%$ higher success rates than the strongest baseline, OpenVLA-OFT~\cite{kim2025fine}. Leveraging intent-level representations, \algacro further enables one-shot skill transfer, achieving $60\%$ higher transfer performance on novel tasks and environments from a single demonstration. Real-robot experiments confirm that \algacro transfers effectively to physical systems, requiring only around 20 demonstrations per task while outperforming the strongest baseline ($\pi_{0.5}$) by $29\%$.

\begin{figure*}
    \centering
    \includegraphics[width=\linewidth]{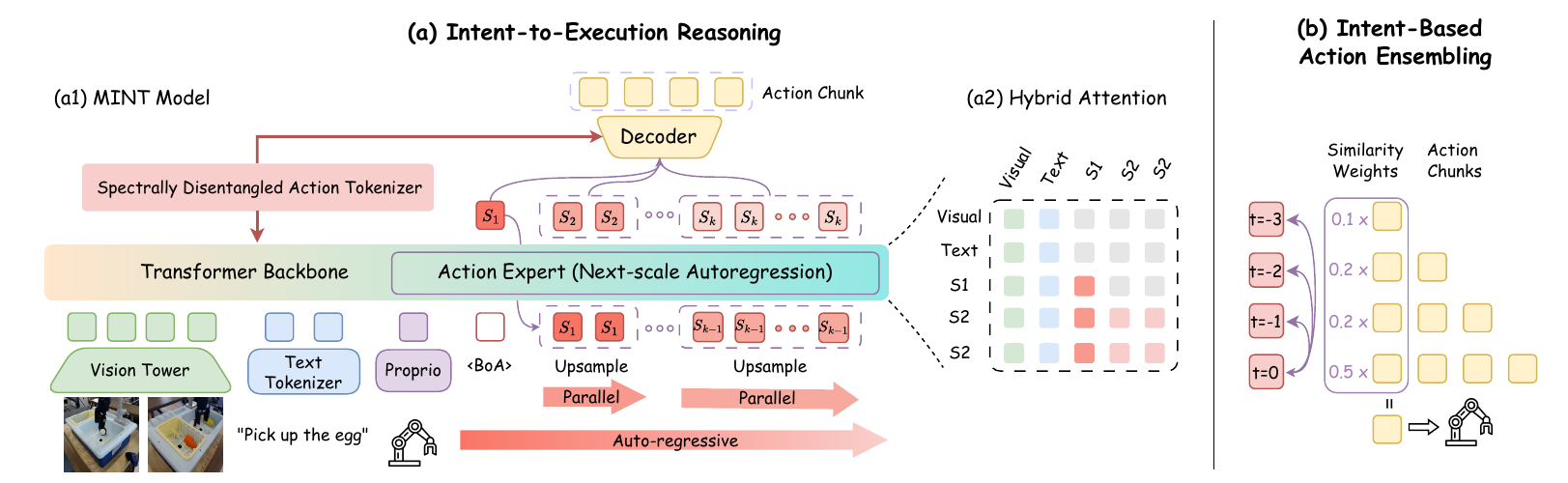}
    \caption{\textbf{\algacro Policy Overview.} (a) \algacro autoregressively predicts action tokens across $K$ temporal scales—moving from a global intent token to high-frequency execution tokens—which are subsequently mapped to continuous trajectories via the decoder. (b) Intent-based action ensemble ensures temporal consistency and smooth behavioral transitions, enhancing stability in long-horizon tasks.}
    \label{fig:overview}
    \vspace{-0.4cm}
\end{figure*}

\section{Related Work}

\subsection{Vision Language Action Models}


The integration of Large Language Models (LLMs)~\cite{touvron2023llama,achiam2023gpt} and Vision-Language Models (VLMs)~\cite{li2023blip,hurst2024gpt} has evolved the prevailing Behavior Cloning paradigm~\cite{levine2016end,chi2025diffusion,ze20243d,zhao2023learning, huang2024diffusion,huang2024goal}, into powerful Vision-Language-Action (VLA) models~\cite{zitkovich2023rt,kim2024openvla,driess2023palm,black2025pi_,black2024pi_0,bjorck2025gr00t,fu2025mergevla,team2024octo}.
However, despite leveraging internet scale pre-training, current VLAs have yet to exhibit the emergent generalization and learning efficiency characteristic of their LLM and VLM counterparts.
We argue that this disparity stems from the fundamental limitation of mimicking raw  trajectories without explicitly comprehending the underlying intent.
Consequently, a framework that can disentangle high-level intent from low-level motion details, while ensuring the learned representations physically executable, is highly desired.



\subsection{Action Tokenization}

Action tokenization~\cite{ye2024latent,bu2025univla,bu2025agibot,schmidt2023learning,chen2025moto,li2025latbot} has emerged as a promising avenue for structuring continuous motor control. 
Mathematical approaches, including direct binning~\cite{brohan2022rt,zhao2023learning} as well as FAST and BEAST~\cite{pertsch2025fast,zhou2025beast}, discretize actions in a structured way and guarantee reconstruction, but they do not enforce explicit constraints to capture behavioral intent. 
Learning based methods, such as VQ-VAE variants~\cite{wang2025vq,lee2024behavior,mete2024quest}, learn tokens automatically and achieve strong compression, but without internal constraints the learned tokens often preserve low-level kinematics rather than intent. 
We address this by constraining tokenization to disentangle intent from execution while keeping actions executable, producing tokens suitable for intent-to-action reasoning.

\subsection{Coarse-to-Fine Tokenization}
Standard Residual VQ methods in VLA~\cite{wang2025vq,liu2025faster} employ a flat hierarchy with uniform capacity across scales, failing to accommodate the inherent asymmetry between sparse, abstract intent and dense, high-frequency execution details. The potential of Multi-Scale VQ is demonstrated by VAR~\cite{tian2024visual} in image generation. CARP~\cite{gong2025carp} mimic this architecture for robotic action chunks. By relying exclusively on time-domain reconstruction over aggregated multi-scale tokens, this design lacks explicit scale wise supervision, leading the hierarchy to prioritize local fidelity over the intent-to-execution structure essential for manipulation. In contrast, we diverge by imposing scale-wise reconstruction constraints explicitly in the frequency domain. This spectral decomposition forces the coarsest scale to exclusively capture global, low-frequency dynamics, ensuring a structural disentanglement of high-level intent from low-level execution details. The intent token opens up two possibilities: intent-based ensembiling, and more crucially, task specification using the intent token, thus one-shot transfer.

\section{Overview}

\algacro is a two-stage imitation learning framework that explicitly disentangles behavioral intent from execution details. It consists of (1) a Spectrally Disentangled Action Tokenizer (\vqvae) that learns structured discrete representations from demonstration trajectories (\figref{fig:teaser}), and (2) \algacro policy that generates actions through progressive intent-to-execution reasoning in the learned token space (\figref{fig:overview}). The \vqvae tokenizer provides a shared action codebook and a decoder, while the \algacro policy learns to predict action tokens in a coarse-to-fine manner and decode them into executable trajectories. Training of \algacro contains two phases:

In the first phase~(\secref{sec:SDAT}), we train \vqvae on demonstration trajectories to obtain multi-scale action representations. Each trajectory is segmented into overlapping action chunks using a sliding window, and each chunk is transformed from the time domain to the frequency domain using the DCT. \vqvae adopts a VQ-VAE~\cite{van2017neural} architecture to learn a discrete action codebook and a quantizer that maps action chunks to tokens.

To induce disentanglement, \vqvae decomposes actions into $K$ temporal scales with progressively increasing capacity (\figref{fig:teaser} Left). The coarsest scale (the \textit{Intent token}) contains a single token intended to capture global, low-frequency structure, while finer scales (the \textit{Execution tokens}) introduce additional tokens that model residual information not explained by coarser ones. All scales share a single codebook. Crucially, the \vqvae tokenizer is trained using progressive reconstruction in the frequency domain: the model is required to reconstruct the frequency-domain trajectory using only the coarsest representation first, then using progressively finer representations, up to all $K$ scales. This design constrains the functionality of tokens at different scales, forcing coarse tokens to explain dominant low-frequency components (\figref{fig:teaser} Right) and finer tokens to specialize in high-frequency residuals. 
Finally, a full reconstruction in the time domain using the union of all scales is applied as an auxiliary objective to ensure faithful recovery of execution details.

In the second phase, we train the \algacro policy that predicts and executes action tokens produced by \vqvae~(\secref{sec:policy}). The policy takes as input the current visual observation, language instruction, and robot proprioceptive state, and outputs an action trajectory. It consists of a vision-language backbone and an action expert. The backbone encodes visual and language inputs using either a standard transformer or a pretrained vision–language model. Conditioned on these features and the robot state, the action expert autoregressively predicts action tokens from coarse to fine scales, generating all tokens within a scale in parallel while maintaining autoregression across scales (\figref{fig:overview} (a)). The predicted tokens are then decoded into continuous trajectories using the decoder inherited from \vqvae. 

We train two variants of \algacro: a language-conditioned version and a language-free version (\algacro-Zero). The former is used to evaluate task performance and robustness on standard manipulation benchmarks, while the latter is designed for one-shot skill transfer. In the transfer setting, an intent token is extracted from a single demonstration and injected into the policy by fixing the coarsest-scale token, while the policy generates execution tokens conditioned on it. We further consider two model scales: a 30M-parameter model adapted from a standard transformer architecture trained from scratch (\algacro-30M), and a 4B-parameter model that combines a pretrained vision–language backbone with a randomly initialized action head (\algacro-4B). Both variants are trained end-to-end in their respective settings.


\section{\vqvaefull}
\label{sec:SDAT}
We propose the \textbf{S}pectrally \textbf{D}isentangled \textbf{A}ction \textbf{T}okenizer (SDAT), a multi-scale framework that explicitly disentangles behavioral intent from low-level execution details.  SDAT introduces a spectral decoder together with a scale-wise spectral reconstruction objective that supervises the frequency composition of actions at different scales, as shown in Algorithm \ref{alg:ms_vqvae}.

\subsection{Action Encoder and Spectrum Decoder}
Let $\mathbf{A} \in \mathbb{R}^{H \times D}$ denote a continuous action sequence of horizon $H$ with action dimension $D$.  
An action encoder $\mathcal{E}$ maps the input sequence into a compressed latent embedding: $f = \mathcal{E}(\mathbf{A}), \quad f \in \mathbb{R}^{L \times C}$, where $L$ denotes the compressed temporal length and $C$ is the latent feature dimension.

Given the latent embedding $f$, a spectrum decoder $\mathcal{D}_{\text{spec}}$ reconstructs the action sequence $\hat{\mathbf{A}} \in \mathbb{R}^{H \times D}$ via an action decoder $\mathcal{D}$ and converts it into frequency-domain representation via the DCT applied along the temporal dimension. For each action dimension $d \in \{1,\dots,D\}$, the DCT coefficients are computed as:
\begin{equation}
    \mathbf{F}_{k,d} = \sum_{h=0}^{H-1} \hat{\mathbf{A}}_{h,d}
    \cos\!\left[\frac{\pi}{H}\left(h+\tfrac{1}{2}\right)k\right],
    k = 0,\dots,H-1,
\end{equation}
where $\mathbf{F} \in \mathbb{R}^{H \times D}$ denotes the resulting frequency-domain representation.  

\subsection{Multi-Scale Residual Quantization}

SDAT utilizes a Multi-Scale Residual Quantization scheme~\cite{tian2024visual,lee2022autoregressive} to decompose the continuous latent embedding $f^{(0)}$ into a multi-scale discrete representation $\mathbf{S} = \{\mathbf{s}_1, \dots, \mathbf{s}_K\}$, where each $\mathbf{s}_k \in \{1, \dots, V\}^{l_k}$ is a discrete token map at resolution $l_k$, representing the quantized features at scale $k$. Let $\mathcal{Z} \in \mathbb{R}^{V \times C}$ denote a shared codebook containing $V$ code vectors, and let $\{l_1, \dots, l_K\}$ be a set of increasing resolutions with $l_K = L$. Quantization is performed recursively on residual features. Let $f^{(k)}$ denote the residual feature at scale $k$. At each scale, the feature is first interpolated to resolution $l_k$ and quantized via $\mathcal{Q}$, producing $\mathbf{s}_k =  \mathcal{Q}(\text{Interpolate}(f^{(k)}, l_k))$. The discrete indices are then mapped to embeddings $\mathbf{z}_k = \text{Lookup}(\mathcal{Z}, \mathbf{s}_k)$. The quantized embeddings $\mathbf{z}_k$ are projected back to the original latent resolution $L$ through an interpolator and a scale-specific projector $\phi_k$, and the residual feature is updated as: $f^{(k+1)} = f^{(k)} - \phi_k(\mathbf{z}_k)$. This forms a coarse-to-fine structure across multiple scales.


\subsection{Scale-wise Spectral Reconstruction}
To enforce spectral disentanglement across quantization scales, we supervise the contribution of each residual level in the frequency domain. Let $\hat{f}^{(k)}$ be the cumulative latent approximation up to scale $k$, formed by summing the quantized residuals: 
\begin{equation}
\hat{f}^{(k)} = \sum_{i=1}^{k} \phi_i\!\left(\text{Lookup}(\mathcal{Z}, \mathbf{s}_i)\right),
\end{equation}
Each cumulative feature $\hat{f}^{(k)}$ is decoded by the shared spectral decoder $\mathcal{D_\text{spec}}$ into a progressively refined action sequence $\hat{\mathbf{A}}^{(k)}$
which is then transformed into the frequency domain as $\mathbf{F}^{(k)} = \text{DCT}(\hat{\mathbf{A}}^{(k)})$. Let the $\mathbf{F} = \text{DCT}(\mathbf{A})$ denote the ground truth, a scale-wise spectral loss enforces consistency between the ground-truth actions and each partial reconstruction:
\begin{equation}
\mathcal{L}_{\text{freq.}}
=
\sum_{k=1}^{K} \lambda_k
\left\|
\mathbf{F}
-
\mathbf{F}^{(k)}
\right\|_{2},
\end{equation}
This encourages early scales to capture low-frequency global structures, while later scales focus on high-frequency details.

\subsection{Training Objective}

The SDAT action tokenizer is trained to capture spectral structure across scales. 
Given an action sequence $\mathbf{A}$, the encoder $\mathcal{E}$ produces a latent $f$, which is discretized via multi-scale residual quantization into $\hat{f}$. 
The spectral decoder $\mathcal{D}_{\text{spec}}$ outputs both frequency domain spectrum $\hat{\mathbf{F}}$ and the reconstructed actions $\hat{\mathbf{A}}$.

The training loss includes the scale-wise spectral reconstruction $\mathcal{L}_{\text{freq.}}$, followed with codebook and commitment losses~\cite{van2017neural}, and a auxiliary $l_1$ reconstruction term. 
Formally, it is defined as:
\begin{equation*}
\mathcal{L} 
= \mathcal{L}_{\text{freq.}}
+ \underbrace{\|\mathrm{sg}(f) - \hat{f}\|_2^2}_{\text{Codebook loss}} + \underbrace{\|f - \mathrm{sg}(\hat{f})\|_2^2}_{\text{Commitment loss}} +
 \alpha \underbrace{\|\mathbf{A} - \hat{\mathbf{A}}\|_1^2}_{\text{Auxiliary loss}},
\end{equation*}
where $\mathrm{sg}(\cdot)$ denotes the stop-gradient operator, and $\alpha$ is a weighting factor.

\begin{algorithm}[t]
\caption{\vqvaefull}
\setstretch{1.1}
\label{alg:ms_vqvae}
\begin{algorithmic}[1]
\STATE \textbf{Inputs:} Action sequence $\mathbf{A}$
\STATE \textbf{Hyperparameters:} scales $K$, resolutions $(l_k)_{k=1}^K$
\STATE \textbf{Initialize:} $f^{(0)} \leftarrow \mathcal{E}(\mathbf{A})$, $\hat{f}^{(0)} \leftarrow 0$, $\mathbf{S} \leftarrow [\,]$, $\mathbf{\mathcal{F}} \leftarrow [\,]$
\FOR{$k = 1, \dots, K$}
    \STATE $\mathbf{s}_k \leftarrow \mathcal{Q}(\text{Interpolate}(f, l_k))$ 
    \STATE $\mathbf{S} \leftarrow \mathbf{S} \cup \{\mathbf{s}_k\}$
    \STATE $\mathbf{z}_k \leftarrow \text{Lookup}(\mathcal{Z}, \mathbf{s}_k)$
    \STATE $\mathbf{z}_k \leftarrow \text{Interpolate}(\mathbf{z}_k, l_K)$ 
    \STATE $f^{(k)} \leftarrow f^{(k-1)} - \phi_k(\mathbf{z}_k)$
    \STATE $\hat{f}^{(k)} \leftarrow \hat{f}^{(k-1)} + \phi_k(\mathbf{z}_k)$ 
    \STATE $\mathbf{F}^{(k)} \leftarrow \mathcal{D}_{\text{spec}}(\hat{f}^{(k)})$
    \STATE $\mathbf{\mathcal{F}} \leftarrow \mathbf{\mathcal{F}} \cup \{\mathbf{F}^{(k)}\}$
\ENDFOR
\STATE $\hat{A} \leftarrow \mathcal{D}(\hat{f}^{(K)})$
\STATE \textbf{Return:} Multi-scale tokens $\mathbf{S}$, frequency domain spectrum $\mathbf{\mathcal{F}}$, reconstruction sequences $\hat{\mathbf{\mathcal{A}}}$
\end{algorithmic}

\end{algorithm}

\section{\algacro Policy Learning}
\label{sec:policy}
The \algacro policy learns \textit{intent-to-execution reasoning} by operating on the multi-scale discrete action tokens produced by SDAT. 
Leveraging \emph{next-scale autoregressive prediction} (\figref{fig:overview} (a1)), the policy performs autoregressive prediction across scales while decoding tokens in parallel within each scale using a hybrid attention mechanism (\figref{fig:overview} (a2)).

\subsection{Next-Scale Autoregressive Modeling}


Building on the multi scale action token maps produced by the SDAT action tokenizer, denoted as $\mathbf{S} = \{\mathbf{s}_1, \dots, \mathbf{s}_K\}$, we model the joint distribution over tokens autoregressively across scales:
\begin{equation}
p(\mathbf{s}_1, \mathbf{s}_2, \dots, \mathbf{s}_K)
=
\prod_{k=1}^{K}
p(\mathbf{s}_k \mid \mathbf{s}_1, \mathbf{s}_2, \dots, \mathbf{s}_{k-1}).
\end{equation}

Each autoregressive unit $\mathbf{s}_k$ is treated as a \emph{token map} rather than a token sequence, and the sequence of coarser-scale token maps $(\mathbf{s}_1, \dots, \mathbf{s}_{k-1})$ serves as the prefix for predicting $\mathbf{s}_k$. At the $k$-th autoregressive step, following \cite{tian2024visual}, all distributions over $l_k$ tokens in $\mathbf{s}_k$ will be generated \emph{in parallel}, conditioned on the prefix token maps $\mathbf{s}_{<k}$ and a scale-specific positional embedding map.

During training, we apply a \emph{hybrid attention mask} to enforce a scale-aware dependency structure, such that the token map at scale $k$ can attend only to token maps from coarser or equal scales $\mathbf{s}_{\leq k}$. The policy is optimized using the standard cross-entropy loss, which measures the discrepancy between the predicted token map $\hat{\mathbf{s}_k}$ and the ground-truth token map $\mathbf{s_k}$ derived from the action sequence.

\subsection{Intent-Based Action Ensemble}

Let $\mathbf{a}_t | \mathbf{o}_t$ denote the predicted action to be executed at time step $t$ conditioned on an observation $\mathbf{o}_t$. The final action at time $t$ is associated with a set of overlapping predictions
$\{\mathbf{a}_t | \mathbf{o}_{t-H}, \ldots, \mathbf{a}_t | \mathbf{o}_{t-1},  \mathbf{a}_t | \mathbf{o}_{t}\}$. During inference, let $\mathbf{s}_1^{(t)} \in \mathbb{R}^C$ denote the intent token associated with the action chunk generated at time step $t$, and $\mathbf{s}_1^{(t-h)}$ denote the intent token of a previous chunk. We derive the final action executed at time $t$ via intent-based action ensemble (\figref{fig:overview} (b)):
\begin{equation}
\mathbf{a}_t
=
\sum_{h=0}^{H}
w_h^{\mathrm{intent}}
\cdot \mathbf{a}_t | \mathbf{o}_{t-h},
\end{equation}
where $w_h^{\mathrm{intent}}$ is an adaptive weight determined by the similarity between behavioral intents. The ensemble weights are computed by measuring the similarity between the current intent token and historical intent tokens:
\begin{equation}
w_h^{\mathrm{intent}} =
\frac{\exp\!\big(\beta \, \langle \mathbf{s}_1^{(t)}, \mathbf{s}_1^{(t-h)} \rangle \big)}
{\sum_{j=0}^{H} \exp\!\big(\beta \, \langle \mathbf{s}_1^{(t)}, \mathbf{s}_1^{(t-j)} \rangle \big)},
\end{equation}
where $\langle \cdot, \cdot \rangle$ denotes the cosine similarity between two intent tokens, 
$\beta > 0$ is a temperature scaling the effect of intent similarity on weight assignment.

Intent based ensemble enabling smooth execution and rapid switching between behaviors. Empirical studies reveal it improves action stability and long-horizon task success.

\subsection{Model Architectures}

%
%

We instantiate our framework with two variant architectures:

\paragraph{MINT-30M} This variant is a lightweight, decoder-only Transformer model trained from scratch, with approximately 30M trainable parameters. Visual inputs are encoded by frozen SigLIP~\cite{zhai2023sigmoid} and DINOv2~\cite{oquab2023dinov2} backbones, while language instructions are processed by a frozen BERT~\cite{devlin2019bert} encoder, and injected into the network via Feature-wise Linear Modulation (FiLM)~\cite{perez2018film} layers. 

\paragraph{MINT-4B} This is a large-scale policy model built upon an existing vision--language architecture used in $\pi_0$ and $\pi_{0.5}$. It employs a PaliGemma-2.6B~\cite{beyer2024paligemma} vision language model together with a SigLIP based visual encoder, both pretrained on large-scale robotic datasets. MINT-4B uses a transformer-based action expert with approximately 300M parameters, which is trained from scratch. The action expert performs next-scale autoregressive prediction over multi-scale action tokens.


\section{Experiments}
\label{sec:exp}
We evaluate our framework on standard benchmarks and the LIBERO-Plus suite to show that explicitly disentangling behavioral intent from execution improves performance and generalization under severe disturbances. We also perform experiments in real-world environments, and we test the one-shot transfer capability of the framework in simulation.

\subsection{Performance Comparison}
\noindent\textbf{Benchmark.} We conduct experiments on three widely adopted robotic manipulation benchmarks: LIBERO, CALVIN, and MetaWorld , which together cover multi-task manipulation, long-horizon compositional reasoning, and task generalization across varying difficulty levels. LIBERO is a simulated benchmark suite composed of five task families: LIBERO-Spatial, LIBERO-Object, LIBERO-Goal, LIBERO-Long and LIBERO-90.  CALVIN features 34 tabletop manipulation tasks across four scene configurations. we evaluate on CALVIN ABCD$\rightarrow$D benchmark, require
policies to follow free-form language instructions and complete 5 tasks in sequence across $500$ different instruction chains. MetaWorld is a large-scale manipulation benchmark consisting of 50 tasks with varying levels of difficulty. Tasks are categorized into Easy, Medium, Hard, and Very Hard groups, reflecting increasing requirements on precision, coordination, and long-horizon control.

\noindent\textbf{Baselines.} We compare against a comprehensive set of baselines across three benchmarks. Specifically, on the LIBERO suite, we compare our method against Diffusion Policy, WorldVLA~\cite{cen2025worldvla}, SmolVLA~\cite{shukor2025smolvla}, both train from scratch, OpenVLA~\cite{kim2024openvla}, OpenVLA-OFT~\cite{kim2025fine}, LAPA~\cite{ye2024latent}, UniVLA~\cite{bu2025univla}, and the $\pi$ family ($\pi_0$~\cite{black2024pi_0}, $\pi_0$-FAST~\cite{pertsch2025fast}, and $\pi_{0.5}$), both pretrain with large robot dataset. For the CALVIN benchmark, we evaluate against RT-1~\cite{brohan2022rt}, RoboVLMs~\cite{li2024towards}, UniVLA, and $\pi_{0.5}$. On Meta-World, we compare against Diffusion Policy, TinyVLA~\cite{wen2025tinyvla}, and $\pi_0$. 

\noindent \textbf{Results.} The results for these experiments are summarized in Table~\ref{tab:performance_results}. MINT consistently matches or surpasses all current state-of-the-art approaches across all reported benchmarks. Notably, our MINT-30M variant surpasses OpenVLA and $\pi_0$ on LIBERO by wide margins despite its significantly smaller size and training from scratch. Our pre-trained MINT-4B variant also outperforms $\pi_{0.5}$ on LIBERO and the $\pi_0$ baseline on Meta-World, where it nearly triples the success rate in the most challenging ``Very Hard'' tasks. On CALVIN, MINT-4B demonstrate superior stability in long-horizon composition. These results demonstrate the versatility of MINT to adapt to diverse task complexities and manipulation settings, showing that MINT not only provides strong performance across scales but also demonstrates robust generalization to difficult, high-precision environments.

\begin{table}[t]
    \centering
    \caption{Performance comparison across LIBERO, CALVIN, and MetaWorld benchmarks}
    \label{tab:performance_results}
    \renewcommand{\arraystretch}{1.2}
    \setlength{\tabcolsep}{4pt}
    \resizebox{\linewidth}{!}{
    \begin{tabular}{l ccccc c}
        \toprule
        
        \multicolumn{7}{c}{\textbf{LIBERO}} \\
        \midrule
        Method & SPATIAL & OBJECT & GOAL & LONG & \textbf{Avg.} & L90 \\
        \midrule
        
        \multicolumn{7}{c}{\textit{Without Pre-training}} \\
        Diffusion Policy~\cite{chi2025diffusion} & 78.3 & 92.5 & 68.3 & 50.5 & 72.4 & -- \\
        MDT~\cite{reuss2024multimodal}            & 78.5 & 87.5 & 73.5 & 64.8 & 76.1 & -- \\
        WorldVLA~\cite{cen2025worldvla}           & 87.6 & 96.2 & 83.4 & 60.0 & 81.8 & -- \\
        SmolVLA~\cite{shukor2025smolvla}           & 93.0 & 94.0 & 91.0 & 77.0 & 88.8 & -- \\
        \rowcolor{oursvivid}
        \textbf{\tf}                                 & \textbf{98.6} & \textbf{99.2} & \textbf{97.4} & \textbf{93.2} & \textbf{97.1} & \textbf{97.4} \\
        
        \midrule
        \multicolumn{7}{c}{\textit{With Pre-training}} \\
        LAPA~\cite{ye2024latent}                  & 73.8 & 74.6 & 58.8 & 55.4 & 65.7 & -- \\
        OpenVLA~\cite{kim2024openvla}             & 84.7 & 88.4 & 79.2 & 53.7 & 76.5 & -- \\
        $\pi_0$-FAST~\cite{pertsch2025fast}       & 96.4 & 96.8 & 88.6 & 60.2 & 85.5 & -- \\
        $\pi_0$~\cite{black2024pi_0}               & 90.0 & 86.0 & 95.0 & 73.0 & 86.0 & -- \\
        UniVLA~\cite{bu2025univla}                & 96.5 & 96.8 & 95.6 & 92.0 & 95.2 & -- \\
        OpenVLA-OFT~\cite{kim2025fine}            & 96.9 & 98.1 & 95.6 & 91.1 & 95.4 & -- \\
        $\pi_{0.5}$~\cite{black2025pi_}            & \textbf{98.8} & 98.2 & 98.0 & 92.4 & 96.9 & 96.0 \\
        \rowcolor{oursvivid}
        \textbf{\vla}                               & 97.4 & \textbf{99.6} & \textbf{98.2} & \textbf{97.8} & \textbf{98.3} & \textbf{98.7} \\
        
        \midrule
        
        \multicolumn{7}{c}{\textbf{CALVIN (ABCD$\rightarrow$D)}} \\
        \midrule
        Method & \multicolumn{5}{c}{Success @ $k$ Tasks} & \textbf{Avg.} \\
        \cmidrule(lr){2-6}
               & 1 & 2 & 3 & 4 & 5 & Len \\
        \midrule
        RT-1~\cite{brohan2022rt}                  & 84.4 & 61.7 & 43.8 & 32.3 & 22.7 & 2.45 \\
        Robo-Flamingo~\cite{li2023vision}         & 96.4 & 89.6 & 82.4 & 74.0 & 66.0 & 4.09 \\
        $\pi_{0.5}$~\cite{black2025pi_}                               & 94.2 & 89.3 & 82.7 & 78.5 & 70.3 & 4.15 \\
        UnifiedVLA~\cite{wang2025unified}              & \textbf{97.9} & \textbf{94.8} & 89.2 & 82.8 & 75.1 & 4.34 \\
        RoboVLMs~\cite{li2024towards}             & 96.7 & 93.0 & 89.9 & 86.5 & 82.6 & 4.49 \\
        \rowcolor{oursvivid}
        \textbf{\vla}                            & 97.4 & 94.2 & \textbf{91.7} & \textbf{88.2} & \textbf{86.1} & \textbf{4.57} \\
        
        \midrule
        
        \multicolumn{7}{c}{\textbf{MetaWorld}} \\
        \midrule
        Method & Easy & Medium & Hard & Very Hard & \textbf{Avg.} & -- \\
        \midrule
        
        Diffusion Policy~\cite{chi2025diffusion}  & 23.1 & 10.7 & 1.9  & 6.1  & 10.5 & -- \\
        TinyVLA~\cite{wen2025tinyvla}              & 77.6 & 21.5 & 11.4 & 15.8 & 31.6 & -- \\
        $\pi_0$~\cite{black2024pi_0}               & 77.9 & 51.8 & 53.3 & 20.0 & 50.8 & -- \\
        \rowcolor{oursvivid}
        \textbf{\vla}                                 & \textbf{82.1} & \textbf{72.4} & \textbf{58.3} & \textbf{56.0} & \textbf{67.2} & -- \\
        
        \bottomrule
    \end{tabular}
    }
    \vspace{-0.5cm}
\end{table}

\subsection{Generalization}
\label{sec:genralization}

\noindent\textbf{Benchmark and Baselines.} We evaluate the robustness of our model against distribution shifts on the LIBERO-PLUS benchmark. This suite assesses seven distinct dimensions of generalization. These dimensions include camera viewpoints involving variations in pose and field of view, robot initial states comprising manipulator pose variations, and language instructions for instruction following tasks. The benchmark also covers light conditions such as intensity and color shifts, background textures reflecting scene appearance changes, sensor noise involving photometric degradation, and object layout encompassing displacement and confounding objects. We compare our models against a wide range of baselines including OpenVLA, UniVLA, $\pi_0$, $\pi_0$-FAST, and $\pi_{0.5}$. We compare MINT+ against $\pi_{0.5}+$, both finetuned on the LIBERO-PLUS dataset.

\noindent\textbf{Result and Analyze.} Table~\ref{tab:robustness_results} demonstrates that MINT exhibits strong and consistent generalization on the LIBERO-PLUS benchmark. Across both model scales, MINT-30M and MINT-4B outperform prior baselines under camera viewpoint and robot initialization shifts, with especially pronounced gains for camera variations. MINT demonstrates stable performance under scene-level variations and sensor perturbations, while preserving reliable instruction following, indicating effective alignment between intent and low-level action execution. Fine-tuning on the LIBERO-PLUS dataset further amplifies these advantages. Compared with $\pi_{0.5}+$, MINT-4B+ leverages the same distributional diversity more effectively, achieving broader and more uniform improvements across perturbation types. This highlights the strength of the MINT framework in generalizing to complex and heterogeneous conditions, rather than relying solely on increased data diversity.

\begin{table}[t]
    \centering
    \caption{Generalization comparison on LIBERO-PLUS }
    \label{tab:robustness_results}
    \renewcommand{\arraystretch}{1.2}
    \setlength{\tabcolsep}{2pt}
    \resizebox{\linewidth}{!}{
        \begin{tabular}{l cccccccc}
            \toprule
            Method & Camera & Robot & Lang. & Light & Back. & Noise & Layout & \textbf{Avg.}  \\
            \midrule
            OpenVLA     & 0.8  & 3.5  & 23.0 & 8.1  & 34.8 & 15.2 & 28.5 & 16.3  \\
            UniVLA      & 1.8  & 46.2 & 69.9 & 69.0 & 81.0 & 21.2 & 31.9 & 45.9  \\
            $\pi_0$     & 13.8 & 6.0  & 58.8 & 85.0 & 81.4 & 79.0 & 68.9 & 56.1  \\
            $\pi_0$-FAST & 65.1 & 21.6 & 61.0 & 73.2 & 73.2 & 74.4 & 68.8 & 62.5  \\
            OpenVLA-OFT & 56.4 & 31.9 & 79.5 & 88.7 & \textbf{93.3} & 75.8 & 74.2 & 71.4 \\
            $\pi_{0.5}$       & 53.0 & \textbf{50.3} & 65.7   & 83.1   & 77.3   & 53.2   & 72.7  & 65.0      \\
            \rowcolor{oursvivid}
            MINT-30M & 61.4 & 41.2 & 61.6   & 92.2 & 77.1 & 76.5 & 76.2 & 69.5    \\
            \rowcolor{oursvivid}
            \textbf{MINT-4B}      & \textbf{72.2} & 42.4 & \textbf{85.8} & \textbf{96.6} & 88.9 & \textbf{90.1} & \textbf{84.6} & \textbf{80.1}  \\
            
            \midrule
            \multicolumn{9}{l}{\textit{Trained with LIBERO Plus}} \\

            OpenVLA-OFT+ & 92.8 & 30.3 & \textbf{85.8} & 94.9 & 93.9 & 89.3 & 77.6 & 80.7 \\

            $\pi_{0.5}$+ & 67.2 & 42.4 & 59.4 & 75.8 & 74.9 & 72.6 & 64.5 & 65.3  \\
            \rowcolor{oursvivid}
            \textbf{MINT-4B+}   & \textbf{95.6} & \textbf{44.6} & 84.7 & \textbf{95.1} & \textbf{94.5} & \textbf{95.2} & \textbf{78.7} & \textbf{84.1}  \\
            \bottomrule
        \end{tabular}
        }
        \vspace{-0.5cm}
\end{table}

\subsection{One-Shot Transfer via Intent Token Injection}
\begin{figure*}[t]
    \centering
    \includegraphics[width=\linewidth]{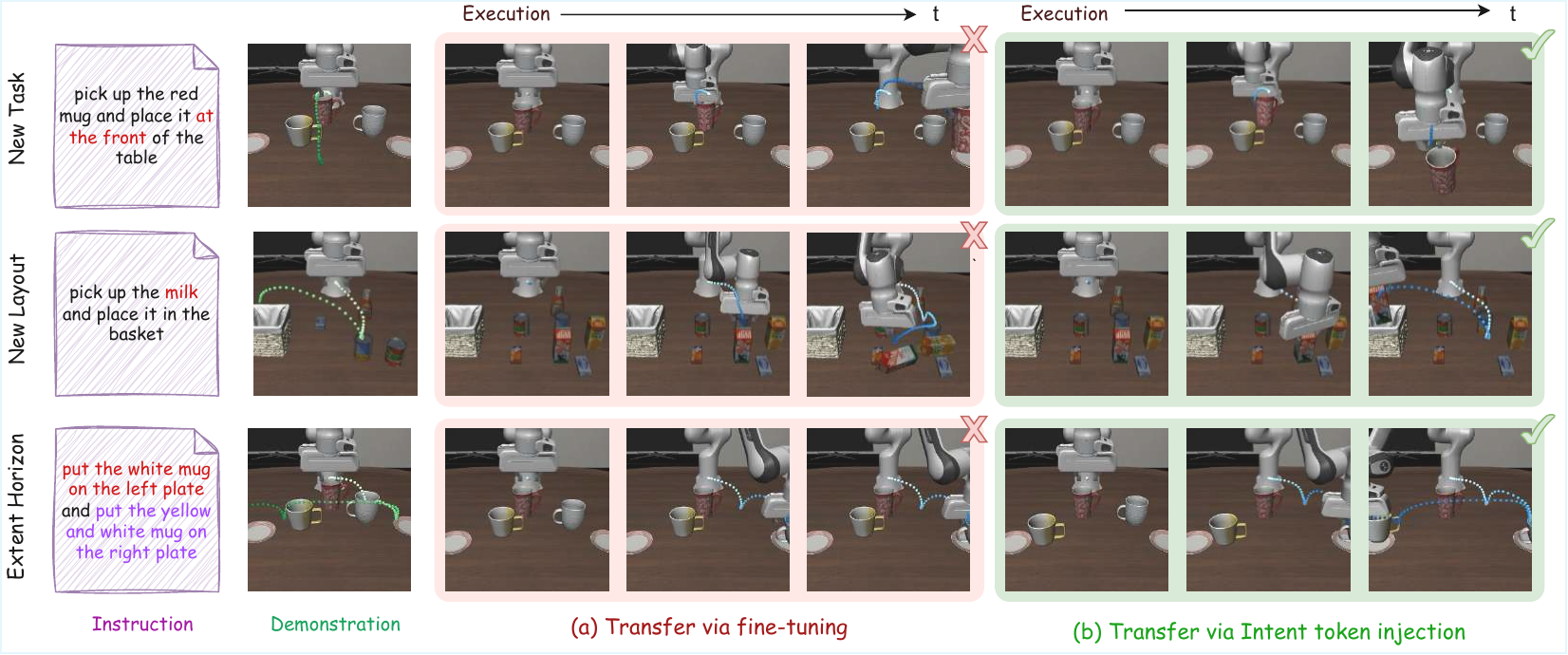}
    \caption{\textbf{One-shot transfer evaluation on OOD tasks in simulation}. We evaluate generalization across three compositional shifts: New Layout, New Task, and Extended Horizon.}
    \label{fig:one-shot}
    \vspace*{-0.2cm}
\end{figure*}


We evaluate one-shot transfer on out-of-distribution tasks by comparing MINT-30M against MINT-Zero-30M. While MINT-30M is one-shot finetuned on a single demonstration for each evaluation task, MINT-Zero-30M conditions on an explicit intent token extracted from a single demonstration via the action tokenizer.
The evaluation focuses on three types of distributional shifts as shown in \figref{fig:one-shot}:
(i) \emph{New Task}, introducing entirely unseen task semantics; 
(ii) \emph{New Layout}, where the task is familiar but the layout is novel; 
(iii) \emph{Extended Horizon}, requiring the execution of longer, sequential actions than observed during training.



We construct a base dataset from LIBERO-90 by excluding the ``\emph{LIVING ROOM SCENE 1-3}'' subset, ensuring that evaluation tasks with novel layouts and unseen semantics remain strictly out-of-distribution. Both MINT-30M and MINT-Zero-30M are trained on this base dataset; for MINT-30M, each evaluation task is further one-shot finetuned using a single demonstration. During inference, MINT-30M uses language-based task specifications, while MINT-Zero-30M is conditioned on the $\mathbf{s}_1$ token extracted from a single demonstration and autoregressively predicts the remaining action tokens via next-scale prediction.



\begin{table}[t]
    \centering
    \caption{One-shot transfer performance comparison.}
    \label{tab:oneshot_transfer}
    \renewcommand{\arraystretch}{1.2}
    \resizebox{\linewidth}{!}{
        \begin{tabular}{l ccccc}
            \toprule
            \makecell[l]{Method} 
            & \makecell[c]{Task\\Specification}
            & \makecell[c]{New\\Task} 
            & \makecell[c]{New\\Layout} 
            & \makecell[c]{Extend\\Horizon} 
            & \textbf{Avg.} \\
            \midrule
            Replay   & Replay   & 0.28 & 0.12 & 0.04 & 0.11 \\
            Fine-tune (MINT-30M)   & Language & 0.42 & 0.08 & 0.00 & 0.17 \\
            \rowcolor{oursvivid}
            \textbf{Intent-injection~(MINT-Zero-30M)}  & Intent   & \textbf{0.90} & \textbf{0.68} & \textbf{0.72} & \textbf{0.77} \\
            \bottomrule
        \end{tabular}
    }
    \vspace{-0.5cm}
\end{table}

\noindent\textbf{Results.} As shown in Table~\ref{tab:oneshot_transfer}, MINT-30M achieves limited one-shot transfer. 
For new layouts, one-shot transfer can cause the model to diverge, while for extended-horizon tasks, one-shot finetuning fails to capture the required behaviors. 
In contrast, intent-based task specification enables effective one-shot transfer, yielding high success rates across new tasks, new layouts, and extended-horizon sequences. 
These results highlight that explicit intent representation provides a more grounded and execution-aligned task specification than language, allowing policies to efficiently transfer to new settings to novel compositions without additional training.

\subsection{Real-World Experiments} 
\begin{figure}[t]
    \centering
    \includegraphics[width=1 \linewidth]{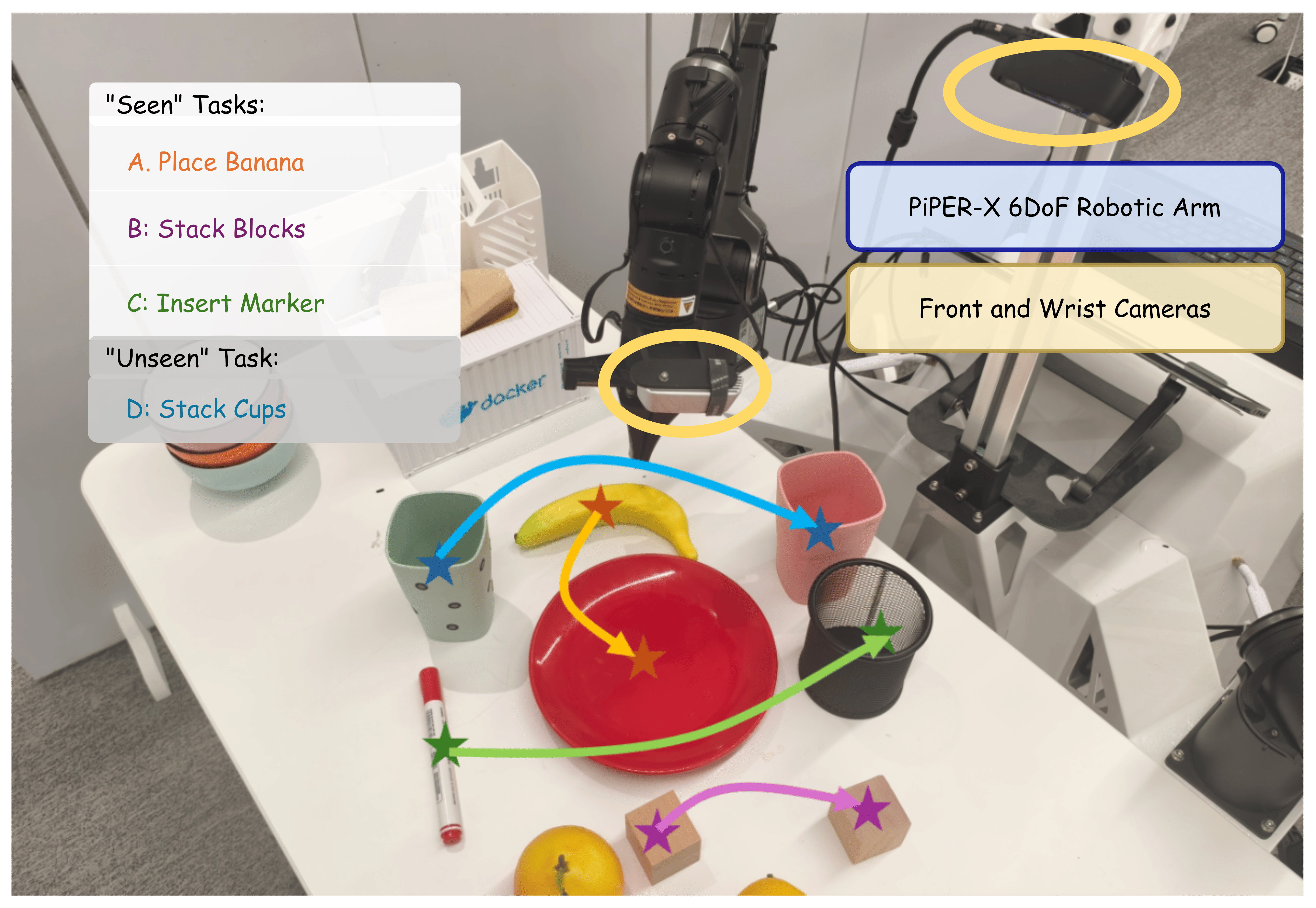}
        \caption{\textbf{Real-world Experiment Setup.}}
    \label{fig:real_world_setup}
    \vspace{-0.5cm}
\end{figure}

We evaluated MINT-4B on four real-world tasks: seen behaviors \textit{(A) Place Banana}, \textit{(B) Stack Blocks}, and \textit{(C) Insert Marker}, alongside an unseen \textit{(D) Stack Cups} task for zero-shot generalization (\figref{fig:real_world_setup}). MINT-4B significantly outperforms all baselines, successfully managing the high-precision coaxial alignment and geometric re-orientation required for structural stability and proper fit (Fig.~\ref{fig: real_world_experiment_results}).  

\begin{figure}[t]
    \centering
    \includegraphics[width=1 \linewidth]{exp_ret.png}
    \caption{\textbf{Real-world task results.}
    The violin plots show Bayesian posterior success rates. The distinct lettering indicates statistically distinguishable policies.}
    \label{fig: real_world_experiment_results}
        \vspace{-0.4cm}
\end{figure}

\noindent\textbf{Training Data.}
The training data consists of a small task-specific dataset and a large-scale prior dataset. 
We use BridgeDataV2~\cite{walke2023bridgedata}, which contains over 60k manipulation trajectories across 24 environments and 13 skills, offering diverse objects, viewpoints, and workspace layouts. 
This dataset is used to pre-train the tokenizer and action head. 
For target tasks, we collect 20 demonstrations per task, totaling 2.4k frames, which are used for in-domain post-training in a multi-task learning setting.

\noindent\textbf{Baselines and Setup.} We evaluated our MINT-4B model against three baselines: a task-specific policy (ACT~\cite{zhao2023learning}), a generalist policy with pretrained weights ($\pi_0$~\cite{black2024pi_0}), and a modified version of the $\pi_{0.5}$~\cite{black2025pi_} policy with a re-initialized action-expert head ($\pi_{0.5}^{*}$). Both MINT-4B and $\pi_{0.5}^{*}$ utilize the same pretrained VLM backbone and were pretrained on BridgeDataV2 before being finetuned on our collected demonstrations. In contrast, ACT was trained individually for each task, and $\pi_0$ was directly finetuned from its pretrained weights. All policies were deployed on a 6-DOF Piper-X robotic arm with dual-camera RGB input, and each policy's performance was evaluated over 20 trials per task to ensure statistical significance.


\noindent\textbf{Results.} 
Using Bayesian posterior analysis, we find MINT statistically distinguishable from all baselines on seen tasks (A) and (B). Notably, MINT significantly outperforms the runner-up $\pi^*_{0.5}$ on (B) \textit{Stack Blocks}, demonstrating superior capability in high-precision axis alignment. Similar performance gaps are observed in the unseen task \textit{(D) Stack Cups}. Despite novel objects, MINT effectively generalizes the shared ``stacking'' intent from task (B), substantially outperforming baselines that overfit to specific object instances.




\subsection{Ablation Studies}
\label{sec:ablation}

To isolate the contributions of the spectral disentanglement objective and the intent-based action ensemble, we conduct ablation studies across CALVIN and LIBERO-LONG baselines.

\begin{figure}[t]
    \centering
    \includegraphics[width=\linewidth]{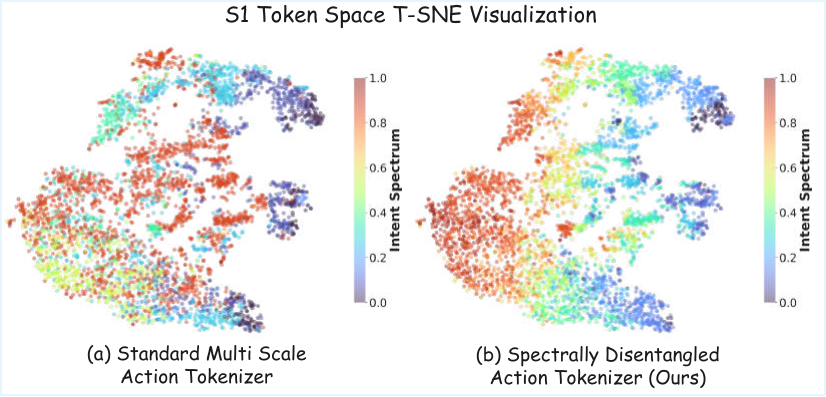}
    \caption{\textbf{Visualization of the Intent Latent Space.} 
    t-SNE of action chunks colored by $\mathbf{s}_1$ tokens (RGB from top-3 PCs). (a) Standard time-domain reconstruction is fragmented. (b) Our SDAT produces coherent
chromatic clusters aligned with action sequences structures.}
    \label{fig:intent_tsne}
    \vspace{-0.2cm}
\end{figure}

\noindent\textbf{Efficacy of Scale-Wise Spectral Decomposition.}
Fig.~\ref{fig:intent_tsne} qualitatively demonstrates that our spectral objective organizes the latent space into coherent behavioral clusters, overcoming the fragmentation observed in standard time-domain reconstruction. Quantitatively (Table~\ref{tab:ablation_unified}), while scale-wise time-domain constraints degrade performance (82.8\%) by overfitting to high-frequency noise, our \textit{Scale-Wise Spectral Loss} yields significant gains (93.4\% on LIBERO-Long, 4.54 length on CALVIN). This confirms that enforcing spectral hierarchy is essential for disentangling global intent from execution details.

\noindent\textbf{Impact of Intent-based Action Ensemble.}
Table~\ref{tab:ablation_unified} shows that our \textit{Intent-Based Action Ensemble} consistently outperforms both Temporal~\cite{zhao2023learning} (89.2\%) and Action-based~\cite{li2024cogact} (90.4\%) baselines. By dynamically modulating aggregation weights based on intent compatibility, our method effectively resolves conflicts during behavioral transitions, achieving the highest success rate (93.2\%) on LIBERO-Long and average sequence length (4.57) on CALVIN.

\begin{table}[t]
    \centering
    \caption{Ablation of training objectives and ensembling.}
    \label{tab:ablation_unified}
    \renewcommand{\arraystretch}{1.2}
    \resizebox{\linewidth}{!}{
    \begin{tabular}{l cc}
        \toprule
        \textbf{Ablation Setting} & \textbf{CALVIN} & \textbf{LIBERO-Long} \\
        \midrule
        
        \multicolumn{3}{l}{\textit{Reconstruction Objectives}} \\

        Terminal Time-Domain Loss & 4.36 & 87.8 \\
        + Terminal Spectral Loss & 4.41 & 88.2 \\
        + Scale-Wise Time-Domain Loss & 4.06 & 82.8 \\
        \rowcolor{oursvivid}
        + Scale-Wise Spectral Loss (Ours) & \textbf{4.54} & \textbf{93.4} \\
        
        \midrule
        
        \multicolumn{3}{l}{\textit{Action Ensemble}} \\

        No Ensemble & 4.09 & 85.8 \\
        Temporal-based Ensemble~\cite{zhao2023learning} & 4.32 & 89.2 \\
        Action-based Ensemble~\cite{li2024cogact} & 4.10 & 90.4 \\
        \rowcolor{oursvivid}
        Intent-based Ensemble (Ours) & \textbf{4.57} & \textbf{93.2} \\
        
        \bottomrule
    \end{tabular}
    }
     \vspace{-0.2cm}
\end{table}

\section{Conclusion} 
\label{sec:conclusion}


We present MINT (\textit{Mimic Intent, Not just Trajectories}), an imitation learning framework that explicitly decouples behavioral intent from low-level execution to address the generalization limitations of current VLA models caused by the entanglement of high-level planning and control dynamics. MINT leverages the Spectrally Disentangled Action Tokenizer (SDAT) to separate low-frequency global intent from high-frequency execution residuals via scale-wise spectral reconstruction, ensuring stable intent representation, improving robustness to environmental variations, achieving state-of-the-art performance across diverse manipulation benchmarks, and enabling effective one-shot skill transfer through intent injection.

\noindent\textbf{Limitation and Future Work.} MINT relies on trajectory demonstrations to learn intent, which limits the diversity of intents to the scope of available datasets. Exploring large-scale network data could provide a richer set of behaviors and broader coverage of tasks, while recombining discrete intent tokens offers a promising avenue for synthesizing novel, long-horizon behaviors zero-shot. Together, these directions could further enhance the generalization and flexibility of intent-driven control.

\bibliographystyle{plainnat}
\bibliography{references}

@article{levine2016end,
  title={End-to-end training of deep visuomotor policies},
  author={Levine, Sergey and Finn, Chelsea and Darrell, Trevor and Abbeel, Pieter},
  journal={Journal of Machine Learning Research},
  volume={17},
  number={39},
  pages={1--40},
  year={2016}
}

@article{chi2025diffusion,
  title={Diffusion policy: Visuomotor policy learning via action diffusion},
  author={Chi, Cheng and Xu, Zhenjia and Feng, Siyuan and Cousineau, Eric and Du, Yilun and Burchfiel, Benjamin and Tedrake, Russ and Song, Shuran},
  journal={The International Journal of Robotics Research},
  volume={44},
  number={10-11},
  pages={1684--1704},
  year={2025},
  publisher={Sage Publications Sage UK: London, England}
}

@article{wang2025unified,
  title={Unified Vision-Language-Action Model},
  author={Wang, Yuqi and Li, Xinghang and Wang, Wenxuan and Zhang, Junbo and Li, Yingyan and Chen, Yuntao and Wang, Xinlong and Zhang, Zhaoxiang},
  journal={arXiv preprint arXiv:2506.19850},
  year={2025}
}

@article{li2024cogact,
  title={Cogact: A foundational vision-language-action model for synergizing cognition and action in robotic manipulation},
  author={Li, Qixiu and Liang, Yaobo and Wang, Zeyu and Luo, Lin and Chen, Xi and Liao, Mozheng and Wei, Fangyun and Deng, Yu and Xu, Sicheng and Zhang, Yizhong and others},
  journal={arXiv preprint arXiv:2411.19650},
  year={2024}
}

@article{ze20243d,
  title={3d diffusion policy: Generalizable visuomotor policy learning via simple 3d representations},
  author={Ze, Yanjie and Zhang, Gu and Zhang, Kangning and Hu, Chenyuan and Wang, Muhan and Xu, Huazhe},
  journal={arXiv preprint arXiv:2403.03954},
  year={2024}
}

@article{touvron2023llama,
  title={Llama 2: Open foundation and fine-tuned chat models},
  author={Touvron, Hugo and Martin, Louis and Stone, Kevin and Albert, Peter and Almahairi, Amjad and Babaei, Yasmine and Bashlykov, Nikolay and Batra, Soumya and Bhargava, Prajjwal and Bhosale, Shruti and others},
  journal={arXiv preprint arXiv:2307.09288},
  year={2023}
}

@article{achiam2023gpt,
  title={Gpt-4 technical report},
  author={Achiam, Josh and Adler, Steven and Agarwal, Sandhini and Ahmad, Lama and Akkaya, Ilge and Aleman, Florencia Leoni and Almeida, Diogo and Altenschmidt, Janko and Altman, Sam and Anadkat, Shyamal and others},
  journal={arXiv preprint arXiv:2303.08774},
  year={2023}
}

@inproceedings{li2023blip,
  title={Blip-2: Bootstrapping language-image pre-training with frozen image encoders and large language models},
  author={Li, Junnan and Li, Dongxu and Savarese, Silvio and Hoi, Steven},
  booktitle={International conference on machine learning},
  pages={19730--19742},
  year={2023},
  organization={PMLR}
}

@article{hurst2024gpt,
  title={Gpt-4o system card},
  author={Hurst, Aaron and Lerer, Adam and Goucher, Adam P and Perelman, Adam and Ramesh, Aditya and Clark, Aidan and Ostrow, AJ and Welihinda, Akila and Hayes, Alan and Radford, Alec and others},
  journal={arXiv preprint arXiv:2410.21276},
  year={2024}
}

@article{driess2023palm,
  title={Palm-e: An embodied multimodal language model},
  author={Driess, Danny and Xia, Fei and Sajjadi, Mehdi SM and Lynch, Corey and Chowdhery, Aakanksha and Wahid, Ayzaan and Tompson, Jonathan and Vuong, Quan and Yu, Tianhe and Huang, Wenlong and others},
  year={2023}
}

@inproceedings{zitkovich2023rt,
  title={Rt-2: Vision-language-action models transfer web knowledge to robotic control},
  author={Zitkovich, Brianna and Yu, Tianhe and Xu, Sichun and Xu, Peng and Xiao, Ted and Xia, Fei and Wu, Jialin and Wohlhart, Paul and Welker, Stefan and Wahid, Ayzaan and others},
  booktitle={Conference on Robot Learning},
  pages={2165--2183},
  year={2023},
  organization={PMLR}
}

@article{kim2024openvla,
  title={Openvla: An open-source vision-language-action model},
  author={Kim, Moo Jin and Pertsch, Karl and Karamcheti, Siddharth and Xiao, Ted and Balakrishna, Ashwin and Nair, Suraj and Rafailov, Rafael and Foster, Ethan and Lam, Grace and Sanketi, Pannag and others},
  journal={arXiv preprint arXiv:2406.09246},
  year={2024}
}

@article{black2024pi_0,
  title={$\pi\_0 $: A Vision-Language-Action Flow Model for General Robot Control},
  author={Black, Kevin and Brown, Noah and Driess, Danny and Esmail, Adnan and Equi, Michael and Finn, Chelsea and Fusai, Niccolo and Groom, Lachy and Hausman, Karol and Ichter, Brian and others},
  journal={arXiv preprint arXiv:2410.24164},
  year={2024}
}

@article{bjorck2025gr00t,
  title={Gr00t n1: An open foundation model for generalist humanoid robots},
  author={Bjorck, Johan and Casta{\~n}eda, Fernando and Cherniadev, Nikita and Da, Xingye and Ding, Runyu and Fan, Linxi and Fang, Yu and Fox, Dieter and Hu, Fengyuan and Huang, Spencer and others},
  journal={arXiv preprint arXiv:2503.14734},
  year={2025}
}

@article{pertsch2025fast,
  title={Fast: Efficient action tokenization for vision-language-action models},
  author={Pertsch, Karl and Stachowicz, Kyle and Ichter, Brian and Driess, Danny and Nair, Suraj and Vuong, Quan and Mees, Oier and Finn, Chelsea and Levine, Sergey},
  journal={arXiv preprint arXiv:2501.09747},
  year={2025}
}

@article{kim2025fine,
  title={Fine-tuning vision-language-action models: Optimizing speed and success},
  author={Kim, Moo Jin and Finn, Chelsea and Liang, Percy},
  journal={arXiv preprint arXiv:2502.19645},
  year={2025}
}

@inproceedings{black2025pi_,
  title={$\pi\_0.5$ : a Vision-Language-Action Model with Open-World Generalization},
  author={Black, Kevin and Brown, Noah and Darpinian, James and Dhabalia, Karan and Driess, Danny and Esmail, Adnan and Equi, Michael Robert and Finn, Chelsea and Fusai, Niccolo and Galliker, Manuel Y and others},
  booktitle={9th Annual Conference on Robot Learning},
  year={2025}
}

@article{van2017neural,
  title={Neural discrete representation learning},
  author={Van Den Oord, Aaron and Vinyals, Oriol and others},
  journal={Advances in neural information processing systems},
  volume={30},
  year={2017}
}

@article{ye2024latent,
  title={Latent action pretraining from videos},
  author={Ye, Seonghyeon and Jang, Joel and Jeon, Byeongguk and Joo, Sejune and Yang, Jianwei and Peng, Baolin and Mandlekar, Ajay and Tan, Reuben and Chao, Yu-Wei and Lin, Bill Yuchen and others},
  journal={arXiv preprint arXiv:2410.11758},
  year={2024}
}

@article{lee2024behavior,
  title={Behavior generation with latent actions},
  author={Lee, Seungjae and Wang, Yibin and Etukuru, Haritheja and Kim, H Jin and Shafiullah, Nur Muhammad Mahi and Pinto, Lerrel},
  journal={arXiv preprint arXiv:2403.03181},
  year={2024}
}

@article{wang2025vq,
  title={VQ-VLA: Improving Vision-Language-Action Models via Scaling Vector-Quantized Action Tokenizers},
  author={Wang, Yating and Zhu, Haoyi and Liu, Mingyu and Yang, Jiange and Fang, Hao-Shu and He, Tong},
  journal={arXiv preprint arXiv:2507.01016},
  year={2025}
}

@article{liu2025faster,
  title={FASTer: Toward Efficient Autoregressive Vision Language Action Modeling via Neural Action Tokenization},
  author={Liu, Yicheng and Zhang, Shiduo and Dong, Zibin and Ye, Baijun and Yuan, Tianyuan and Yu, Xiaopeng and Yin, Linqi and Lu, Chenhao and Shi, Junhao and Yu, Luca Jiang-Tao and others},
  journal={arXiv preprint arXiv:2512.04952},
  year={2025}
}

@article{li2025latbot,
  title={LatBot: Distilling Universal Latent Actions for Vision-Language-Action Models},
  author={Li, Zuolei and Gao, Xingyu and Wang, Xiaofan and Fu, Jianlong},
  journal={arXiv preprint arXiv:2511.23034},
  year={2025}
}

@inproceedings{chen2025moto,
  title={Moto: Latent motion token as the bridging language for learning robot manipulation from videos},
  author={Chen, Yi and Ge, Yuying and Tang, Weiliang and Li, Yizhuo and Ge, Yixiao and Ding, Mingyu and Shan, Ying and Liu, Xihui},
  booktitle={Proceedings of the IEEE/CVF International Conference on Computer Vision},
  pages={19752--19763},
  year={2025}
}

@article{bu2025univla,
  title={Univla: Learning to act anywhere with task-centric latent actions},
  author={Bu, Qingwen and Yang, Yanting and Cai, Jisong and Gao, Shenyuan and Ren, Guanghui and Yao, Maoqing and Luo, Ping and Li, Hongyang},
  journal={arXiv preprint arXiv:2505.06111},
  year={2025}
}

@article{tian2024visual,
  title={Visual autoregressive modeling: Scalable image generation via next-scale prediction},
  author={Tian, Keyu and Jiang, Yi and Yuan, Zehuan and Peng, Bingyue and Wang, Liwei},
  journal={Advances in neural information processing systems},
  volume={37},
  pages={84839--84865},
  year={2024}
}

@inproceedings{gong2025carp,
  title={Carp: Visuomotor policy learning via coarse-to-fine autoregressive prediction},
  author={Gong, Zhefei and Ding, Pengxiang and Lyu, Shangke and Huang, Siteng and Sun, Mingyang and Zhao, Wei and Fan, Zhaoxin and Wang, Donglin},
  booktitle={Proceedings of the IEEE/CVF International Conference on Computer Vision},
  pages={13460--13470},
  year={2025}
}

@article{liu2023libero,
  title={Libero: Benchmarking knowledge transfer for lifelong robot learning},
  author={Liu, Bo and Zhu, Yifeng and Gao, Chongkai and Feng, Yihao and Liu, Qiang and Zhu, Yuke and Stone, Peter},
  journal={Advances in Neural Information Processing Systems},
  volume={36},
  pages={44776--44791},
  year={2023}
}

@inproceedings{yu2020meta,
  title={Meta-world: A benchmark and evaluation for multi-task and meta reinforcement learning},
  author={Yu, Tianhe and Quillen, Deirdre and He, Zhanpeng and Julian, Ryan and Hausman, Karol and Finn, Chelsea and Levine, Sergey},
  booktitle={Conference on robot learning},
  pages={1094--1100},
  year={2020},
  organization={PMLR}
}

@article{mees2022calvin,
  title={Calvin: A benchmark for language-conditioned policy learning for long-horizon robot manipulation tasks},
  author={Mees, Oier and Hermann, Lukas and Rosete-Beas, Erick and Burgard, Wolfram},
  journal={IEEE Robotics and Automation Letters},
  volume={7},
  number={3},
  pages={7327--7334},
  year={2022},
  publisher={IEEE}
}

@article{mete2024quest,
  title={Quest: Self-supervised skill abstractions for learning continuous control},
  author={Mete, Atharva and Xue, Haotian and Wilcox, Albert and Chen, Yongxin and Garg, Animesh},
  journal={Advances in Neural Information Processing Systems},
  volume={37},
  pages={4062--4089},
  year={2024}
}

@article{fei2025libero,
  title={Libero-plus: In-depth robustness analysis of vision-language-action models},
  author={Fei, Senyu and Wang, Siyin and Shi, Junhao and Dai, Zihao and Cai, Jikun and Qian, Pengfang and Ji, Li and He, Xinzhe and Zhang, Shiduo and Fei, Zhaoye and others},
  journal={arXiv preprint arXiv:2510.13626},
  year={2025}
}

@article{reuss2024multimodal,
  title={Multimodal diffusion transformer: Learning versatile behavior from multimodal goals},
  author={Reuss, Moritz and Ya{\u{g}}murlu, {\"O}mer Erdin{\c{c}} and Wenzel, Fabian and Lioutikov, Rudolf},
  journal={arXiv preprint arXiv:2407.05996},
  year={2024}
}

@article{cen2025worldvla,
  title={WorldVLA: Towards Autoregressive Action World Model},
  author={Cen, Jun and Yu, Chaohui and Yuan, Hangjie and Jiang, Yuming and Huang, Siteng and Guo, Jiayan and Li, Xin and Song, Yibing and Luo, Hao and Wang, Fan and others},
  journal={arXiv preprint arXiv:2506.21539},
  year={2025}
}

@article{shukor2025smolvla,
  title={Smolvla: A vision-language-action model for affordable and efficient robotics},
  author={Shukor, Mustafa and Aubakirova, Dana and Capuano, Francesco and Kooijmans, Pepijn and Palma, Steven and Zouitine, Adil and Aractingi, Michel and Pascal, Caroline and Russi, Martino and Marafioti, Andres and others},
  journal={arXiv preprint arXiv:2506.01844},
  year={2025}
}

@article{team2024octo,
  title={Octo: An open-source generalist robot policy},
  author={Team, Octo Model and Ghosh, Dibya and Walke, Homer and Pertsch, Karl and Black, Kevin and Mees, Oier and Dasari, Sudeep and Hejna, Joey and Kreiman, Tobias and Xu, Charles and others},
  journal={arXiv preprint arXiv:2405.12213},
  year={2024}
}

@article{wen2025tinyvla,
  title={Tinyvla: Towards fast, data-efficient vision-language-action models for robotic manipulation},
  author={Wen, Junjie and Zhu, Yichen and Li, Jinming and Zhu, Minjie and Tang, Zhibin and Wu, Kun and Xu, Zhiyuan and Liu, Ning and Cheng, Ran and Shen, Chaomin and others},
  journal={IEEE Robotics and Automation Letters},
  year={2025},
  publisher={IEEE}
}

@article{brohan2022rt,
  title={Rt-1: Robotics transformer for real-world control at scale},
  author={Brohan, Anthony and Brown, Noah and Carbajal, Justice and Chebotar, Yevgen and Dabis, Joseph and Finn, Chelsea and Gopalakrishnan, Keerthana and Hausman, Karol and Herzog, Alex and Hsu, Jasmine and others},
  journal={arXiv preprint arXiv:2212.06817},
  year={2022}
}

@article{li2023vision,
  title={Vision-language foundation models as effective robot imitators},
  author={Li, Xinghang and Liu, Minghuan and Zhang, Hanbo and Yu, Cunjun and Xu, Jie and Wu, Hongtao and Cheang, Chilam and Jing, Ya and Zhang, Weinan and Liu, Huaping and others},
  journal={arXiv preprint arXiv:2311.01378},
  year={2023}
}

@article{li2024towards,
  title={Towards Generalist Robot Policies: What Matters in Building Vision-Language-Action Models},
  author={Li, Xinghang and Li, Peiyan and Liu, Minghuan and Wang, Dong and Liu, Jirong and Kang, Bingyi and Ma, Xiao and Kong, Tao and Zhang, Hanbo and Liu, Huaping},
  journal={arXiv preprint arXiv:2412.14058},
  year={2024}
}

@article{fu2025mergevla,
  title={MergeVLA: Cross-Skill Model Merging Toward a Generalist Vision-Language-Action Agent},
  author={Fu, Yuxia and Zhang, Zhizhen and Zhang, Yuqi and Wang, Zijian and Huang, Zi and Luo, Yadan},
  journal={arXiv preprint arXiv:2511.18810},
  year={2025}
}

@article{ahmed2006discrete,
  title={Discrete cosine transform},
  author={Ahmed, Nasir and Natarajan, T\_ and Rao, Kamisetty R},
  journal={IEEE transactions on Computers},
  volume={100},
  number={1},
  pages={90--93},
  year={2006},
  publisher={IEEE}
}

@article{zhou2025beast,
  title={BEAST: Efficient Tokenization of B-Splines Encoded Action Sequences for Imitation Learning},
  author={Zhou, Hongyi and Liao, Weiran and Huang, Xi and Tang, Yucheng and Otto, Fabian and Jia, Xiaogang and Jiang, Xinkai and Hilber, Simon and Li, Ge and Wang, Qian and others},
  journal={arXiv preprint arXiv:2506.06072},
  year={2025}
}

@article{bu2025agibot,
  title={Agibot world colosseo: A large-scale manipulation platform for scalable and intelligent embodied systems},
  author={Bu, Qingwen and Cai, Jisong and Chen, Li and Cui, Xiuqi and Ding, Yan and Feng, Siyuan and Gao, Shenyuan and He, Xindong and Hu, Xuan and Huang, Xu and others},
  journal={arXiv preprint arXiv:2503.06669},
  year={2025}
}

@article{schmidt2023learning,
  title={Learning to act without actions},
  author={Schmidt, Dominik and Jiang, Minqi},
  journal={arXiv preprint arXiv:2312.10812},
  year={2023}
}

@inproceedings{zhai2023sigmoid,
  title={Sigmoid loss for language image pre-training},
  author={Zhai, Xiaohua and Mustafa, Basil and Kolesnikov, Alexander and Beyer, Lucas},
  booktitle={Proceedings of the IEEE/CVF international conference on computer vision},
  pages={11975--11986},
  year={2023}
}

@article{oquab2023dinov2,
  title={Dinov2: Learning robust visual features without supervision},
  author={Oquab, Maxime and Darcet, Timoth{\'e}e and Moutakanni, Th{\'e}o and Vo, Huy and Szafraniec, Marc and Khalidov, Vasil and Fernandez, Pierre and Haziza, Daniel and Massa, Francisco and El-Nouby, Alaaeldin and others},
  journal={arXiv preprint arXiv:2304.07193},
  year={2023}
}

@inproceedings{devlin2019bert,
  title={Bert: Pre-training of deep bidirectional transformers for language understanding},
  author={Devlin, Jacob and Chang, Ming-Wei and Lee, Kenton and Toutanova, Kristina},
  booktitle={Proceedings of the 2019 conference of the North American chapter of the association for computational linguistics: human language technologies, volume 1 (long and short papers)},
  pages={4171--4186},
  year={2019}
}

@inproceedings{perez2018film,
  title={Film: Visual reasoning with a general conditioning layer},
  author={Perez, Ethan and Strub, Florian and De Vries, Harm and Dumoulin, Vincent and Courville, Aaron},
  booktitle={Proceedings of the AAAI conference on artificial intelligence},
  volume={32},
  number={1},
  year={2018}
}

@article{beyer2024paligemma,
  title={Paligemma: A versatile 3b vlm for transfer},
  author={Beyer, Lucas and Steiner, Andreas and Pinto, Andr{\'e} Susano and Kolesnikov, Alexander and Wang, Xiao and Salz, Daniel and Neumann, Maxim and Alabdulmohsin, Ibrahim and Tschannen, Michael and Bugliarello, Emanuele and others},
  journal={arXiv preprint arXiv:2407.07726},
  year={2024}
}

@inproceedings{lee2022autoregressive,
  title={Autoregressive image generation using residual quantization},
  author={Lee, Doyup and Kim, Chiheon and Kim, Saehoon and Cho, Minsu and Han, Wook-Shin},
  booktitle={Proceedings of the IEEE/CVF conference on computer vision and pattern recognition},
  pages={11523--11532},
  year={2022}
}

@inproceedings{huang2024diffusion,
  title={Diffusion models as optimizers for efficient planning in offline rl},
  author={Huang, Renming and Pei, Yunqiang and Wang, Guoqing and Zhang, Yangming and Yang, Yang and Wang, Peng and Shen, Hengtao},
  booktitle={European Conference on Computer Vision},
  pages={1--17},
  year={2024},
  organization={Springer}
}

@article{huang2024goal,
  title={Goal-Reaching Policy Learning from Non-Expert Observations via Effective Subgoal Guidance},
  author={Huang, RenMing and Liu, Shaochong and Pei, Yunqiang and Wang, Peng and Wang, Guoqing and Yang, Yang and Shen, Hengtao},
  journal={arXiv preprint arXiv:2409.03996},
  year={2024}
}

@inproceedings{walke2023bridgedata,
  title={Bridgedata v2: A dataset for robot learning at scale},
  author={Walke, Homer Rich and Black, Kevin and Zhao, Tony Z and Vuong, Quan and Zheng, Chongyi and Hansen-Estruch, Philippe and He, Andre Wang and Myers, Vivek and Kim, Moo Jin and Du, Max and others},
  booktitle={Conference on Robot Learning},
  pages={1723--1736},
  year={2023},
  organization={PMLR}
}

@article{zhao2023learning,
  title={Learning fine-grained bimanual manipulation with low-cost hardware},
  author={Zhao, Tony Z and Kumar, Vikash and Levine, Sergey and Finn, Chelsea},
  journal={arXiv preprint arXiv:2304.13705},
  year={2023}
}

\clearpage
\appendix

\subsection{Implementation Details}
All experiments in this work are conducted with distributed
training on $4 \times$ NVIDIA H200 GPUs.

\subsubsection{SDAT}
we employ 1D CNN architectures for the action encoder and spectrum decoder. To ensure physical consistency across heterogeneous action spaces, we initially project translation, rotation, and gripper states via separate MLPs, processing them with Group CNNs in early layers to extract modality-specific features before fusing them for joint encoding. We utilize Exponential Moving Average (EMA) for codebook updates to effectively prevent codebook collapse and ensure stable discrete latent space learning. Furthermore, we explicitly exclude the binary gripper dimension from the Discrete Cosine Transform (DCT) and spectral reconstruction. 

\subsubsection{MINT-30M}
MINT-30M is a lightweight baseline architecture without a VLM backbone, designed to evaluate the effectiveness of our framework when trained entirely from scratch. The model contains approximately 30M trainable parameters. For language processing, we employ BERT~\cite{devlin2019bert} to encode the language command $l_t$. Since MINT-30M does not rely on a large language model backbone, the encoded language features are injected into the policy network via FiLM conditioning, enabling effective language-controlled behavior and improving multi-task generalization. Visual observations are encoded using a frozen, pre-trained Vision Transformer (ViT), following the practice of~\cite{kim2024openvla}. Specifically, we combine features from SigLIP and DINOv2 through feature concatenation to leverage complementary visual representations. The action expert shares the same backbone parameters as the policy network, without introducing an additional model component. During inference, MINT-30M follows a decoder-only Transformer formulation. Although we adopt a scale wise decoding strategy, the model remains compatible with KV caching, allowing efficient autoregressive inference.

\subsubsection{MINT-4B}
MINT-4B follows the overall design philosophy of $\pi_{0.5}$~\cite{black2025pi_} and is built upon the PaliGemma VLM backbone. PaliGemma combines a SigLIP vision encoder with the Gemma-2B language model, and employs multi-query attention with the following configuration: width $=2048$, depth $=18$, MLP dimension $=16{,}384$, number of attention heads $=18$, number of key--value heads $=1$, and head dimension $=256$. Following $\pi_{0.5}$, we adopt a lightweight Transformer as the action expert with reduced capacity (width $=1024$, MLP dimension $=4096$), resulting in approximately 300M parameters. Unlike $\pi_{0.5}$, which uses a DiT-based architecture for flow matching, we formulate the action expert as a decoder-only Transformer. This design choice enables direct compatibility with our scale-wise autoregressive decoding strategy. We initialize the VLM backbone using the publicly released $\pi_{0.5}$ pre-trained parameters, which were trained on large-scale robotic datasets. In contrast, the action expert is randomly initialized and trained from scratch within our framework.

\subsubsection{Hyperparameter Details}
To facilitate reproducibility, we detail the training hyperparameters of all components in Table~\ref{tab:SDAT_config} and Table~\ref{tab:MINT_configs}. Specifically, The SigLIP-based visual encoder contains $400$ million parameters, and DINOv2-based visual encoder contains $300$ million parameters.

\begin{table}[t]
\centering
\caption{Training recipes for SDAT across different benchmarks.}
\label{tab:SDAT_config}
\renewcommand{\arraystretch}{1.15}
\resizebox{\columnwidth}{!}{%
\begin{tabular}{lcccc}
\midrule
\textbf{Parameter} & \textbf{LIBERO} & \textbf{CALVIN} & \textbf{MetaWorld} & \textbf{BridgeV2} \\
\midrule
Codebook Size        & 512  & 512  & 256  & 1024 \\
Code Dim  & 32   & 32   & 32   & 64   \\
Action Horizon       & 16   & 32   & 16   & 16   \\
Scales               & \textnormal{[1,2,4]} & \textnormal{[1,2,3,4]} & \textnormal{[1,2,4]} & \textnormal{[1,2,3,4]} \\
\midrule
Optimizer            & \multicolumn{4}{c}{AdamW} \\
Batch Size           & \multicolumn{4}{c}{1024} \\
Learning Rate        & \multicolumn{4}{c}{3.0$\times10^{-5}$} \\
EMA Ratio            & \multicolumn{4}{c}{0.99} \\
Weight Decay         & \multicolumn{4}{c}{0.01} \\
Optimizer Momentum   & \multicolumn{4}{c}{$\beta_1=0.9,\ \beta_2=0.95$} \\
\bottomrule
\end{tabular}
}
\end{table}

\begin{table}[t]
\centering
\caption{Training recipes for MINT-30M and MINT-4B.}
\label{tab:MINT_configs}
\renewcommand{\arraystretch}{1.15}
\begin{tabular}{lc}
\toprule
\multicolumn{2}{c}{\textbf{MINT-30M}} \\
\midrule
Vision Encoder        & SigLIP + DINOv2 \\
Language Encoder      & BERT \\
Transformer Layers    & 8 \\
Attention Heads       & 12 \\
MLP Dim               & 1024 \\
Width                 & 256 \\
\midrule
Optimizer             & AdamW \\
Batch Size            & 128 \\
Learning Rate         & 3.0$\times10^{-4}$ \\
Weight Decay          & 0.01 \\
Optimizer Momentum    & $\beta_1=0.9,\ \beta_2=0.95$ \\
\bottomrule
\end{tabular}

\begin{tabular}{lc}
\toprule
\multicolumn{2}{c}{\textbf{MINT-4B}} \\
\midrule
Vision Encoder        & SigLIP \\
LLM Backbone          & Gemma-2B \\
Action Expert         & Gemma-300M \\
\midrule
Optimizer             & AdamW \\
Batch Size            & 128 \\
Learning Rate         & 2.0$\times10^{-4}$ \\
Weight Decay          & 0.01 \\
Optimizer Momentum    & $\beta_1=0.9,\ \beta_2=0.95$ \\
\bottomrule
\end{tabular}
\end{table}

\begin{table*}[t]
\centering
\caption{Learning efficiency comparison measured by success rate at different training iterations}
\label{tab:learning_efficient}
\renewcommand{\arraystretch}{1.2}
\setlength{\tabcolsep}{6pt}
\begin{tabular}{lcccccccccc}
\toprule
\textbf{Method}  & $1k$  & $2k$  & $3k$  & $4k$  & $5k$  & $6k$  & $7k$  & $8k$  & $9k$  & $10k$ \\
\midrule

\multicolumn{11}{c}{\textit{Without Pre-training}} \\

ACT
& 0.06 & 0.21 & 0.27 & 0.43 & 0.53 & 0.58 & 0.61 & 0.67 & 0.65 & 0.65 \\
\rowcolor{oursvivid}
MINT-30M 
& 0.00 & 0.43 & 0.74 & 0.84 & 0.87 & 0.86 & 0.92 & 0.92 & 0.93 & 0.95 \\
\midrule

\multicolumn{11}{c}{\textit{With Pre-training}} \\

$\pi_0$-FAST 
& 0.35 & 0.55 & 0.67 & 0.78 & 0.76 & 0.84 & 0.81 & 0.85 & 0.84 & 0.83 \\
$\pi_{0.5}$ 
& 0.39 & 0.64 & 0.73 & 0.78 & 0.80 & 0.81 & 0.84 & 0.82 & 0.85 & 0.89 \\
\rowcolor{oursvivid}
MINT-4B 
& 0.53 & 0.76 & 0.82 & 0.90 & 0.94 & 0.96 & 0.95 & 0.96 & 0.97 & 0.97 \\

\bottomrule
\end{tabular}
\end{table*}

\begin{table}[t]
    \centering
    \caption{Ablation study on the number of scales and action chunk horizon.}
    \label{tab:ablation_results_scale_horizon}
    \renewcommand{\arraystretch}{1.3}
    \setlength{\tabcolsep}{10pt}
    \begin{tabular}{l c c}
        \toprule
        \multicolumn{3}{c}{\textbf{Number of Scales Ablation}} \\
        \midrule
        \makecell[l]{Num of\\Scales} & \makecell[c]{CALVIN\\AVG. LEN} & \makecell[c]{LIBERO-LONG\\Success Rate(\%)} \\
        \midrule
        \textnormal{(1)}          & 2.12  & 42.8  \\
        \textnormal{(1,4)}        & 4.06  & 78.4  \\
        \textnormal{(1,2,4)}      & 4.46  & \textbf{93.6}  \\
        \textnormal{(1,2,3,4)}    & \textbf{4.57}  & 92.2  \\
        \textnormal{(1,2,4,6,8)}  & 4.32  & 88.6  \\
        \midrule
        \multicolumn{3}{c}{\textbf{Action Chunk Horizon Ablation}} \\
        \midrule
        \makecell[l]{Chunk\\Horizon} & \makecell[c]{CALVIN\\AVG. LEN} & \makecell[c]{LIBERO-LONG\\Success Rate(\%)} \\
        \midrule
        8   & 3.74  & 80.6  \\
        16  & 4.47  & \textbf{93.2}  \\
        32  & \textbf{4.49}  & 86.6  \\
        64  & 4.26  & 87.4  \\
        \bottomrule
    \end{tabular}
\end{table}

\subsection{Evaluation Details}
We describe all evaluation tasks and training datasets used in our experiments. We detail the distribution of initial conditions and scoring criteria.

\noindent\textbf{Libero Benchmark}. We follow the training and evaluation setup of~\citet{liu2023libero}. We evaluate on the Libero-Spatial, Libero-Object, Libero-Goal and Libero-Long benchmarking suites and use the corresponding datasets provided by the authors for training. We combine all datasets into one dataset with $270k$ samples, and train one policy jointly on all. We train all policies for a total of $30k$ iterations ($\approx$ 15 epochs). We use the re-rendered datasets of~\citet{kim2024openvla} for our experiments. Success is evaluated as a binary criterion per episode.


\noindent\textbf{CLAVIN}. We follow the standard training and evaluation protocol of the CLAVIN ABCD$\rightarrow$D benchmark, a language-conditioned robotic manipulation dataset consisting of $24k$ human-teleoperated demonstration trajectories. Each trajectory spans up to $64$ timesteps and covers $34$ predefined primitive skills, including object manipulation, drawer interaction, and button or switch control. The dataset is divided into four splits (A, B, C, D). Policies are trained on splits A--C and evaluated on split D. During evaluation, an agent is required to execute a sequence of $5$ randomly sampled tasks in order. We perform $500$ evaluation rollouts on split D, reporting both the success rate of completing the full task sequence and the average number of successfully completed tasks per episode. The Franka Emika Panda robot is controlled in Delta End-Effector space with a discrete gripper, and observations include both static and wrist-mounted RGB cameras. All policies are trained for a total of $30k$ iterations ($\approx 5$ epochs).


\noindent\textbf{Meta-World Benchmark}. We evaluate our method on the Meta-World benchmark, which comprises $50$ diverse robotic manipulation tasks designed for multi-task learning and generalization evaluation. Each task includes multiple variations with randomized initial object states and goal configurations. For each task, we use the demonstration dataset provided by \textit{LeRobot}, which contains $50$ high-quality trajectories per task collected under the standard Meta-World observation and action interfaces, resulting in a total of $2{,}500$ demonstrations across all tasks. Demonstrations are generated with randomized initial conditions to ensure sufficient intra-task diversity. All tasks are jointly used for training a single policy. Policies are trained for $5k$ iterations ($\approx 6$ epochs) and evaluated using the standard Meta-World success criteria.

\noindent\textbf{LIBERO-Plus Benchmark}. We additionally evaluate our method on the LIBERO-Plus benchmark~\cite{fei2025libero}, which is explicitly designed to assess generalization performance under a diverse set of controlled perturbations. LIBERO-Plus extends the original LIBERO benchmark by systematically introducing variations along multiple factors that are critical for evaluating robustness and generalization in language-conditioned robotic manipulation. LIBERO-Plus comprises a total of 10{,}030 tasks spanning seven perturbation factors, each targeting a distinct source of distribution shift. Specifically, the benchmark includes: (1) \emph{Object Layout} perturbations, which introduce confounding objects and displace target objects; (2) \emph{Camera Viewpoint} variations, including changes in camera position, orientation, and field of view; (3) \emph{Robot Initial State} perturbations that vary the manipulator’s initial pose; (4) \emph{Language Instruction} perturbations generated via LLM-based instruction rewriting; (5) \emph{Lighting Conditions}, covering variations in light intensity, direction, color, and shadowing; (6) \emph{Background Texture} changes that alter scene and surface appearance; and (7) \emph{Sensor Noise}, which introduces photometric distortions and image degradation. We follow the evaluation protocol provided by the benchmark and report success as a binary criterion per episode.

\noindent\textbf{Real-World Benchmark}. 
We evaluate real-world performance on a 6-DoF Piper-X robotic arm equipped with a parallel gripper. The benchmark consists of four manipulation tasks designed to assess language-conditioned control, real-time perception, and generalization under varying physical configurations. The benchmark includes four tasks, each paired with a fixed natural language instruction:
\begin{enumerate}
    \item \textit{Place Banana}: grasping a banana from varying initial positions and placing it onto a plate with varying position and color (``place the banana on the plate'').
    
    \item \textit{Stack Blocks}: grasping a block from varying initial positions and stacking it onto another block placed at different locations (``stack the right block on the left block'').
    
    \item \textit{Insert Marker}: picking up a red marker pen from varying initial positions, rotating it to the correct orientation, and inserting it into a black holder with varying poses (``insert the red marker pen into the black holder'').
    
    \item \textit{Stack Cups (Zero-Shot)}: grasping a green cup from varying initial positions and stacking it onto a pink cup placed at different locations, which is used exclusively for zero-shot evaluation (``stack the green cup on the pink cup'').
\end{enumerate}

For the first three tasks, we collect 20 demonstration trajectories per task using teleoperated manipulation. During data collection, the environment is randomly configured to introduce variations in object color and position. Demonstrations are recorded at 10~Hz with a horizon of 90 frames, resulting in a total of 5.4K real-world samples. No demonstrations are collected for the zero-shot cup-stacking task. For ACT, we train a separate policy for each task. In contrast, for $\pi_0$, $\pi_{0.5}$, and our method MINT-4B, a single policy is fine-tuned jointly across all available real-world tasks. Evaluation is conducted along three dimensions: performance on the limited real-world training set, generalization to unseen environment configurations, and zero-shot performance on the unseen cup-stacking task.

\begin{figure*}
    \centering
    \includegraphics[width=0.9\linewidth]{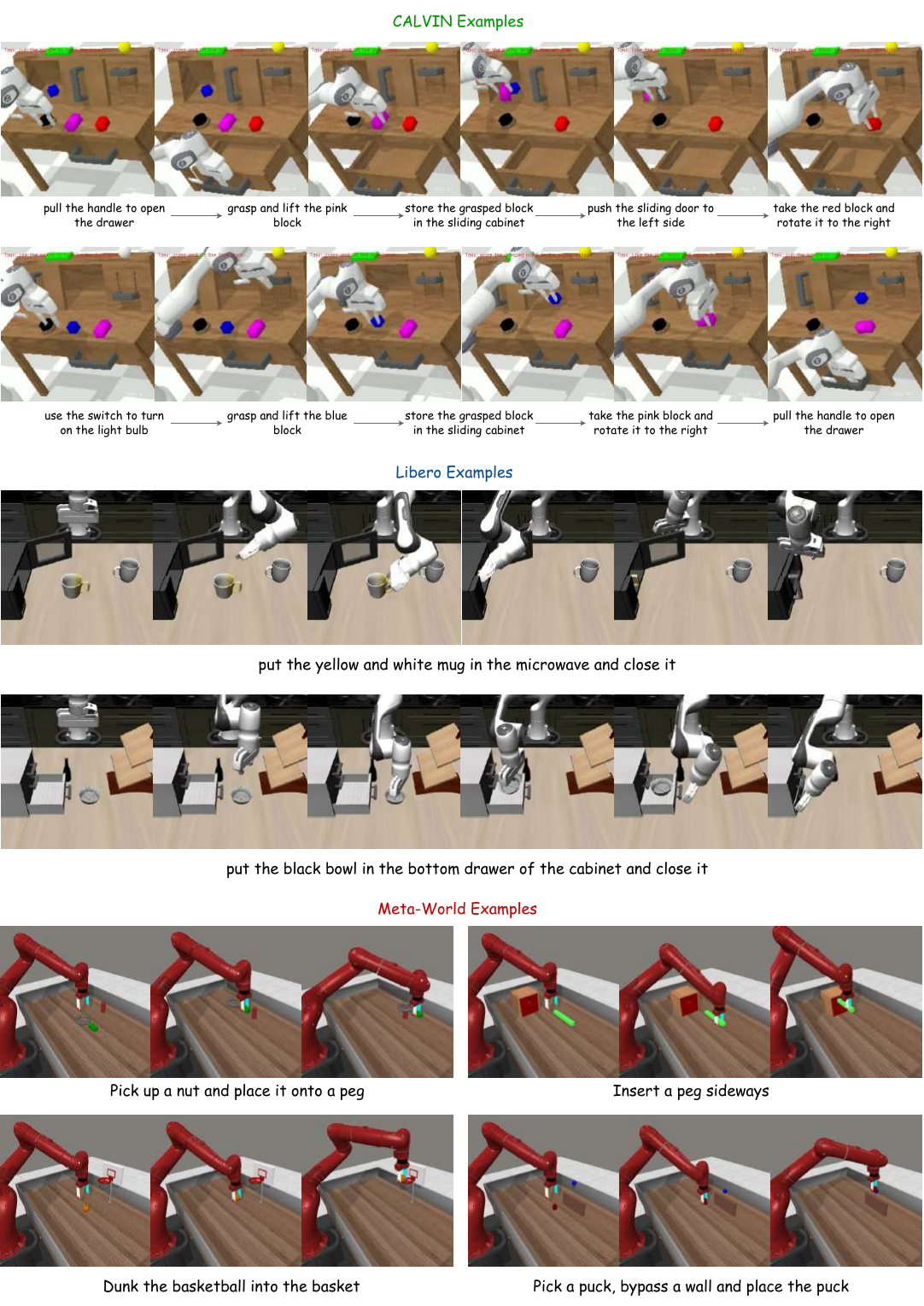}
    \caption{Visualization of MINT-4B on CALVIN, Libero, Meta-World Benchmarks.}
    \label{fig:performace}
\end{figure*}

\begin{figure*}
    \centering
    \includegraphics[width=0.9\linewidth]{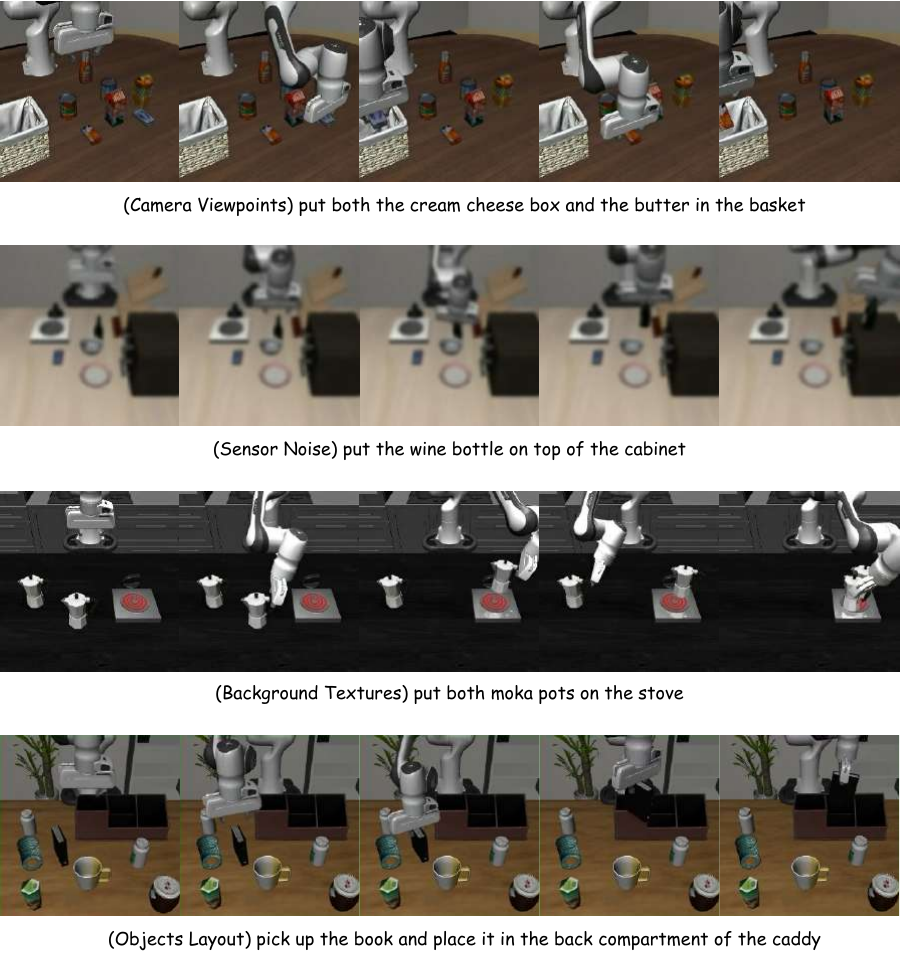}
    \caption{Visualization of MINT-4B on Libero-Plus.}
    \label{fig:libero_plus_examples}
\end{figure*}

\begin{figure*}
    \centering
    \includegraphics[width=0.9\linewidth]{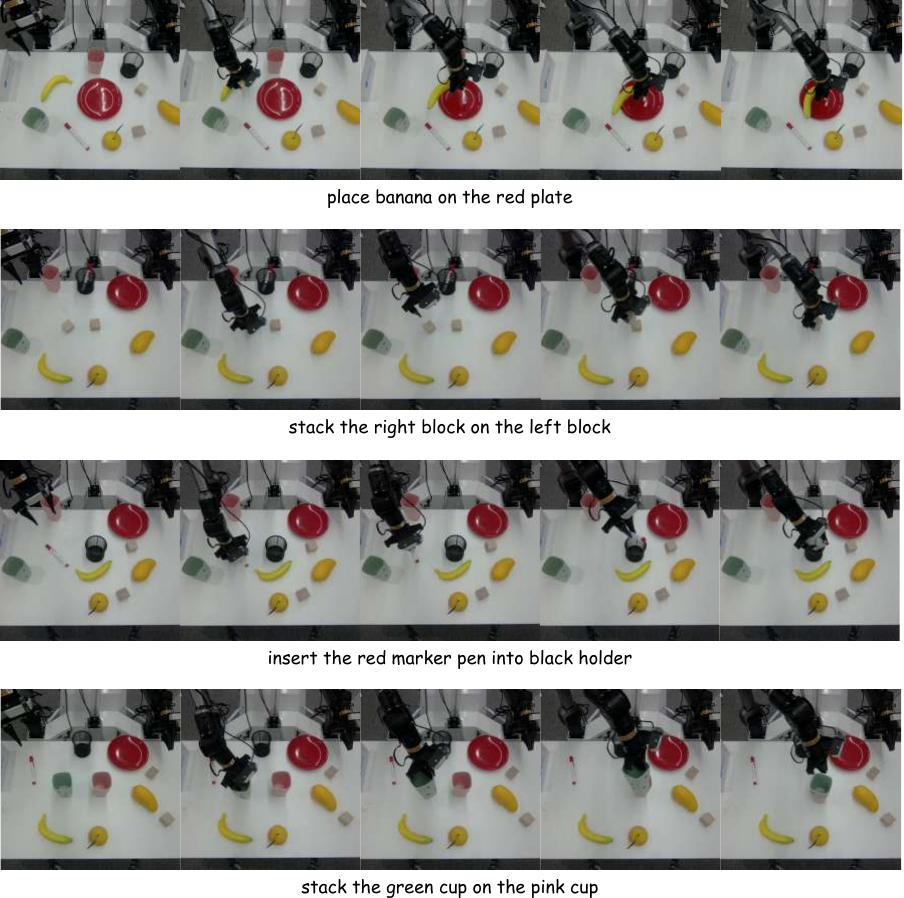}
    \caption{Visualization of MINT-4B on Real-World Tasks}
    \label{fig:real_examples}
\end{figure*}

\noindent\textbf{Execution Examples}. We provide qualitative execution examples across simulated and real-world benchmarks to illustrate the behavioral characteristics of the learned policies. As shown in~\figref{fig:performace}, on CALVIN, the policy successfully executes long sequences of compositional tasks, demonstrating reliable task transitions and sustained performance over extended horizons. On LIBERO and Meta-World, the policy exhibits precise object manipulation and consistent goal completion across diverse task configurations. For LIBERO-Plus, execution examples~(\figref{fig:libero_plus_examples}) highlight robustness under substantial visual, linguistic, and physical perturbations, including changes in camera viewpoints, lighting conditions, background textures, and object layouts. Despite these distribution shifts, the policy maintains stable control and task completion behavior. Real-world execution examples~(\figref{fig:real_examples}) demonstrate that the learned policy transfers effectively to physical robotic systems. The robot performs object placement, stacking, and insertion tasks with accurate perception-action coordination, and successfully completes a zero-shot cup-stacking task without additional demonstrations. These results qualitatively validate the generalization and robustness claims supported by the quantitative evaluations.

\begin{table}[t]
    \centering
    \caption{Performance comparison across LIBERO, CALVIN, and MetaWorld benchmarks}
    \label{tab:performance_results}
    \renewcommand{\arraystretch}{1.2}
    \setlength{\tabcolsep}{4pt}
    \resizebox{\linewidth}{!}{
    \begin{tabular}{l ccccc c}
        \toprule
        
        \multicolumn{7}{c}{\textbf{LIBERO}} \\
        \midrule
        Method & SPATIAL & OBJECT & GOAL & LONG & \textbf{Avg.} & L90 \\
        \midrule
        
        \multicolumn{7}{c}{\textit{Without Pre-training}} \\
        Diffusion Policy~\cite{chi2025diffusion} & 78.3 & 92.5 & 68.3 & 50.5 & 72.4 & -- \\
        MDT~\cite{reuss2024multimodal}            & 78.5 & 87.5 & 73.5 & 64.8 & 76.1 & -- \\
        WorldVLA~\cite{cen2025worldvla}           & 87.6 & 96.2 & 83.4 & 60.0 & 81.8 & -- \\
        SmolVLA~\cite{shukor2025smolvla}           & 93.0 & 94.0 & 91.0 & 77.0 & 88.8 & -- \\
        \rowcolor{oursvivid}
        MINT-30M                       & 98.6 & 99.2 & 97.4 & 93.2 & 97.1 & 97.4 \\
        
        \midrule
        \multicolumn{7}{c}{\textit{With Pre-training}} \\
        LAPA~\cite{ye2024latent}                  & 73.8 & 74.6 & 58.8 & 55.4 & 65.7 & -- \\
        VLACache                                  & 83.8 & 85.8 & 76.4 & 52.8 & 74.7 & -- \\
        Octo                                      & 78.9 & 85.7 & 84.6 & 51.1 & 75.1 & -- \\
        OpenVLA~\cite{kim2024openvla}             & 84.7 & 88.4 & 79.2 & 53.7 & 76.5 & -- \\
        MAIL                                      & 74.3 & 90.1 & 81.8 & 78.6 & 81.2 & -- \\
        DiT Policy                                & 84.2 & 96.3 & 85.4 & 63.8 & 82.4 & -- \\
        CoT-VLA                                   & 87.5 & 91.6 & 87.6 & 69.0 & 83.9 & -- \\
        Think-Act                                 & 88.0 & 91.0 & 87.0 & 71.0 & 84.3 & -- \\
        $\pi_0$-FAST~\cite{pertsch2025fast}       & 96.4 & 96.8 & 88.6 & 60.2 & 85.5 & -- \\
        $\pi_0$~\cite{black2024pi_0}               & 90.0 & 86.0 & 95.0 & 73.0 & 86.0 & -- \\
        UniVLA~\cite{bu2025univla}                & 96.5 & 96.8 & 95.6 & 92.0 & 95.2 & -- \\
        OpenVLA-OFT~\cite{kim2025fine}            & 96.9 & 98.1 & 95.6 & 91.1 & 95.4 & -- \\
        MemoryVLA                                 & 98.4 & 98.4 & 96.4 & 93.4 & 96.7 & 95.6 \\
        $\pi_{0.5}$~\cite{black2025pi_}            & 98.8 & 98.2 & 98.0 & 92.4 & 96.9 & 96.0 \\
        FlowerVLA                                 & 97.5 & 99.1 & 96.1 & 94.9 & 96.9 & 94.7 \\
        \rowcolor{oursvivid}
        MINT-4B                           & 97.4 & 99.6 & 98.2 & 97.8 & 98.3 & 98.7 \\
        
        \midrule
        
        \multicolumn{7}{c}{\textbf{CALVIN (ABCD$\rightarrow$D)}} \\
        \midrule
        Method & \multicolumn{5}{c}{Success @ $k$ Tasks (\%)} & Avg. Len \\
        \cmidrule(lr){2-6}
               & 1 & 2 & 3 & 4 & 5 &  \\
        \midrule
        MCIL                                      & 37.3 & 2.7 & 0.2 & 0.0 & 0.0 & 0.40 \\
        RT-1~\cite{brohan2022rt}                  & 84.4 & 61.7 & 43.8 & 32.3 & 22.7 & 2.45 \\
        Robo-Flamingo~\cite{li2023vision}         & 96.4 & 89.6 & 82.4 & 74.0 & 66.0 & 4.08 \\
        GR-1                                      & 94.9 & 89.6 & 84.4 & 78.9 & 73.1 & 4.21 \\
        ReconVLA                                  & 98.0 & 90.0 & 84.5 & 78.5 & 70.5 & 4.22 \\
        UniVLA                                    & 94.8 & 90.6 & 86.2 & 83.4 & 69.0 & 4.24 \\
        UP-VLA                                    & 96.2 & 92.1 & 87.9 & 84.2 & 81.2 & 4.42 \\
        RoboVLMs~\cite{li2024towards}             & 96.7 & 93.0 & 89.9 & 86.5 & 82.6 & 4.49 \\
        MDT                                       & 98.6 & 95.8 & 91.6 & 86.2 & 80.1 & 4.52 \\
        \rowcolor{oursvivid}
        MINT-4B                                      & 97.4 & 94.2 & 91.7 & 88.2 & 86.1 & 4.58 \\
        
        \midrule
        
        \multicolumn{7}{c}{\textbf{MetaWorld}} \\
        \midrule
        Method & Easy & Medium & Hard & Very Hard & \textbf{Avg.} & -- \\
        \midrule
        Diffusion Policy~\cite{chi2025diffusion}  & 23.1 & 10.7 & 1.9  & 6.1  & 10.5 & -- \\
        TinyVLA~\cite{wen2025tinyvla}              & 77.6 & 21.5 & 11.4 & 15.8 & 31.6 & -- \\
        $\pi_0$~\cite{black2024pi_0}               & 77.9 & 51.8 & 53.3 & 20.0 & 50.8 & -- \\
        \rowcolor{oursvivid}
        MINT-4B                                      & 82.1 & 72.4 & 58.3 & 56.0 & 67.2 & -- \\
        
        \bottomrule
    \end{tabular}
    }
\end{table}

\subsection{More Ablation Studies}
\subsubsection{Learning Efficiency}
We evaluate learning efficiency by measuring success rates at different training iterations. Results in Table~\ref{tab:learning_efficient} show that our approach converges significantly faster than baseline methods, demonstrating improved data efficiency. Notably, even without pre-training, the lightweight MINT-30M model achieves rapid performance gains.

\subsubsection{Ablation on Action Horizon and Number of Scales}
We ablate both the number of spectral scales and the action chunk horizon to analyze their impact on performance. Results in Table~\ref{tab:ablation_results_scale_horizon} indicate that moderate multi-scale configurations provide the best trade-off between expressiveness and optimization stability. Similarly, intermediate action horizons balance long-term planning capability with prediction accuracy, while excessively long horizons degrade performance due to increased modeling difficulty.

\subsection{Additional Results}
\subsubsection{Reconstruction accuracy analysis}
\begin{figure*}
    \centering
    \includegraphics[width=\linewidth]{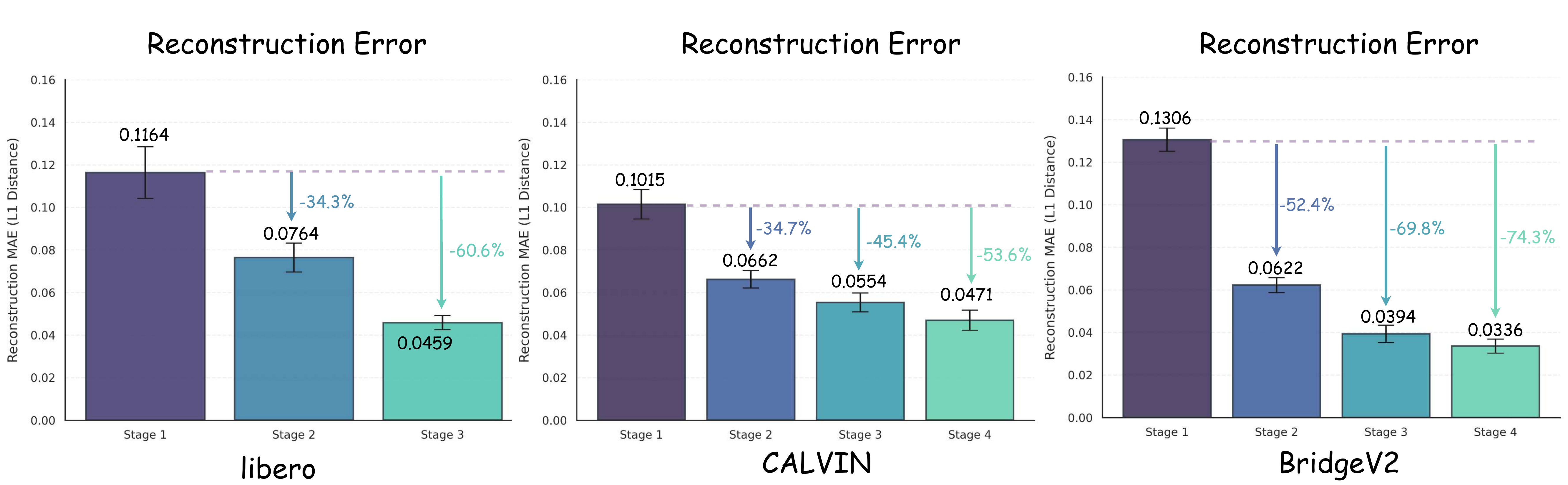}
    \caption{Multi-scale reconstruction error decreases as the number of spectral scales increases}
    \label{fig:multi_scale_recon}
\end{figure*}

\begin{figure*}
    \centering
    \includegraphics[width=\linewidth]{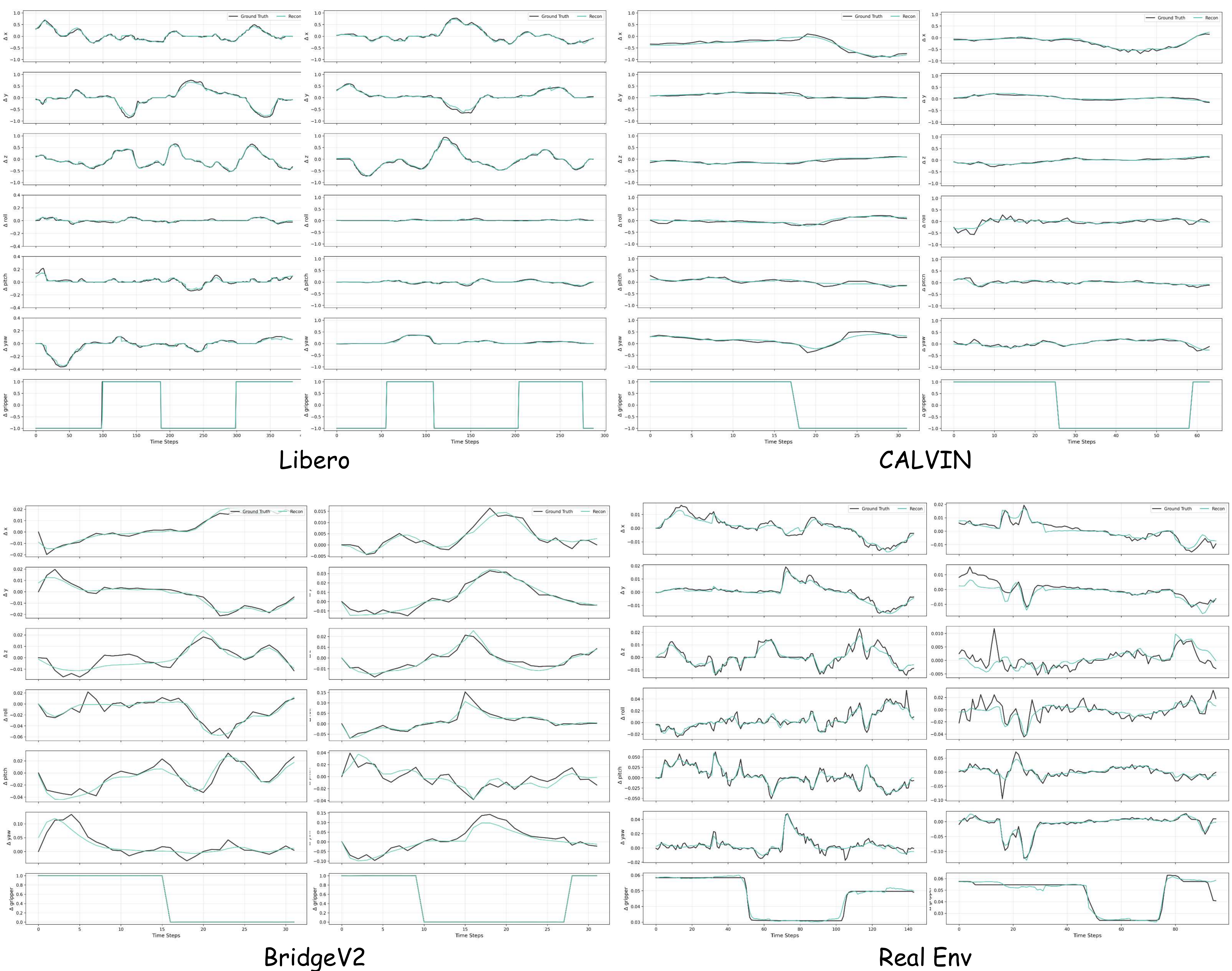}
    \caption{Visualization of action reconstruction results on representative trajectories from the Libero, CALVIN, Bridge and Real-Env datasets.}
    \label{fig:recon_visual}
\end{figure*}

We analyze reconstruction accuracy from two complementary perspectives. We study how the multiple spectral scales affects reconstruction fidelity~(\figref{fig:multi_scale_recon}). Second, we qualitatively compare reconstructed action trajectories against ground-truth trajectories to assess execution-level accuracy~(\figref{fig:recon_visual}). To evaluate the effect of multi-scale decomposition, we measure reconstruction error as the number of spectral scales increases.  We further visualize reconstructed action trajectories alongside their corresponding ground-truth trajectories. In these visualizations, reconstructed trajectories closely follow the temporal structure of the original actions, with deviations primarily occurring in high-frequency regions corresponding to fine-grained motion adjustments. This behavior suggests that the learned representation preserves the global structure of the trajectory while allowing flexible modeling of execution-level variations. These analyses demonstrate that the proposed multi-scale spectral representation improves reconstruction accuracy and yields faithful trajectory reconstructions, supporting its suitability as a structured action representation for downstream policy learning.

\subsubsection{Intent token analysis}
\begin{figure*}
    \centering
    \includegraphics[width=\linewidth]{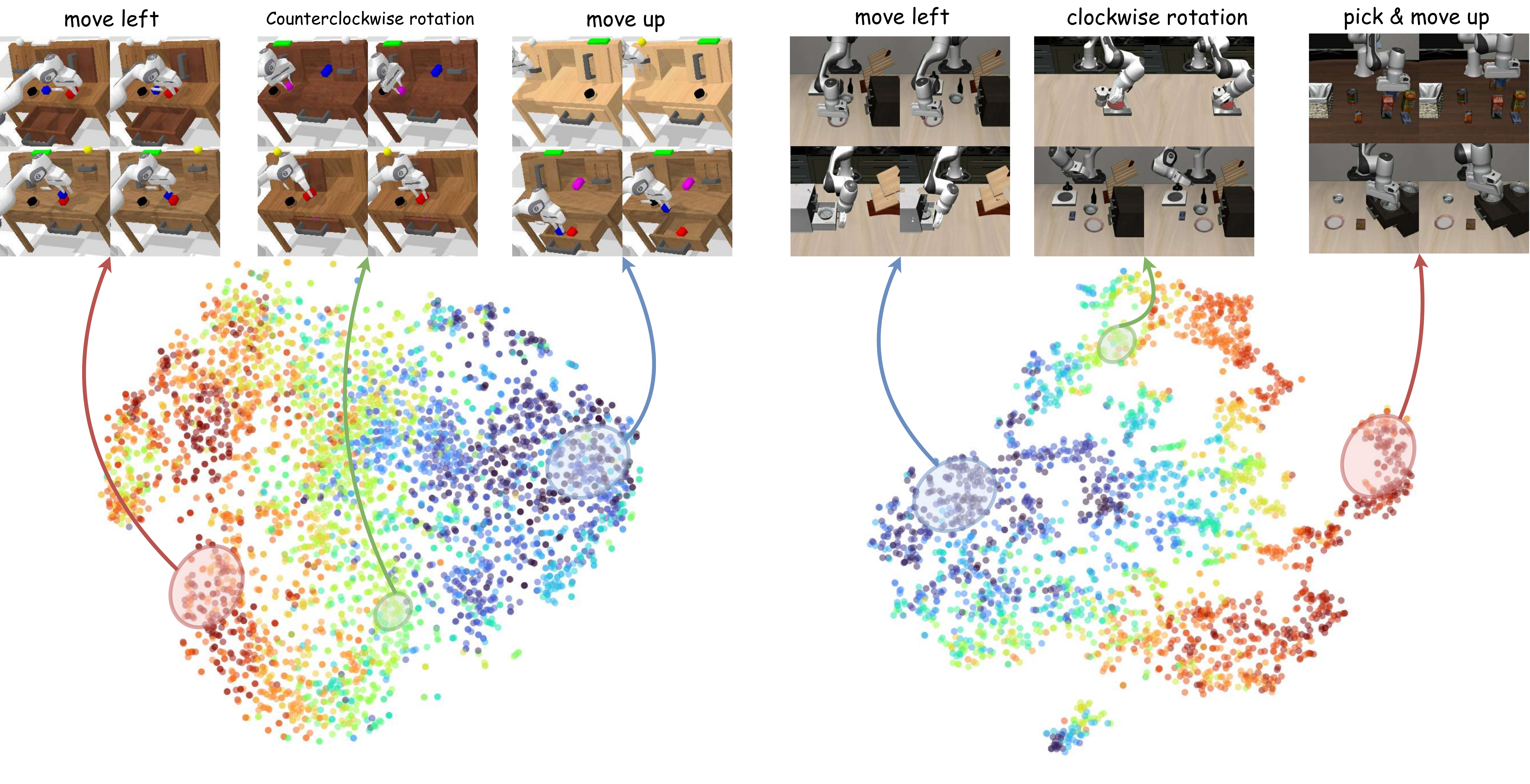}
    \caption{\textbf{Visualization of the Intent Latent Space.} 
    t-SNE of action chunks colored by $\mathbf{s}_1$ tokens. Demonstrate that learned tokens form distinct clusters corresponding to semantic behaviors across LIBERO and CALVIN.}
    \label{fig:intent_examples}
\end{figure*}

In addition to the analysis presented in the main paper, we provide extended intent token visualizations on both the LIBERO and CALVIN benchmarks~(\figref{fig:intent_examples}). We project the learned low-frequency intent tokens (S1 tokens) into a two-dimensional space using t-SNE to examine their semantic structure.

Consistent with the observations in the main text, the S1 token space exhibits clear clustering patterns corresponding to semantically coherent behaviors, such as object pickup, forward motion, and rotational manipulation. Importantly, these clusters remain stable across different tasks, indicating that the learned intent tokens capture task-level behavioral abstractions rather than dataset-specific artifacts. Similar clustering behavior is observed on CALVIN, despite its longer task horizons and sequential structure, suggesting that the disentanglement between intent and execution generalizes across benchmarks with distinct temporal and compositional characteristics.

\subsubsection{Additional Performance Results}
In Table~\ref{tab:performance_results}, We report full benchmark results across LIBERO, CALVIN, and Meta-World to provide a comprehensive comparison against both pre-trained and non-pre-trained baselines. Results demonstrate that our method consistently achieves strong performance across all benchmarks, with particularly notable gains in long-horizon and multi-task settings.

\subsubsection{Additional Libero-PLUS Results}
We provide two complementary tables to offer a more comprehensive analysis on the LIBERO-Plus benchmark. The Table~\ref{tab:robustness_results} reports performance across an expanded set of baselines, enabling a direct comparison with a wide range of prior approaches under identical perturbation settings. The Table~\ref{tab:libero_plus_breakdown} presents a finer-grained breakdown of performance across different LIBERO suites (Spatial, Object, Goal, and Long) and perturbation types. 
\begin{table*}[t]
    \centering
    \caption{Generalization comparison on LIBERO-PLUS }
    \label{tab:robustness_results}
    \renewcommand{\arraystretch}{1.2}
    \setlength{\tabcolsep}{6pt}
    \resizebox{\linewidth}{!}{
        \begin{tabular}{l cccccccc}
            \toprule
            \makecell[l]{Method} 
            & \makecell[c]{Camera\\Viewpoints} 
            & \makecell[c]{Robot\\Initial States} 
            & \makecell[c]{Language\\Instructions} 
            & \makecell[c]{Light\\Conditions} 
            & \makecell[c]{Background\\Textures} 
            & \makecell[c]{Sensor\\Noise} 
            & \makecell[c]{Objects\\Layout} 
            & \textbf{Avg.}  \\
            \midrule
            OpenVLA     & 0.8  & 3.5  & 23.0 & 8.1  & 34.8 & 15.2 & 28.5 & 16.3  \\
            NORA        & 2.2  & 37.0 & 65.1 & 45.7 & 58.6 & 12.8 & 62.1 & 40.5 \\
            WorldVLA    & 0.1  & 27.9 & 41.6 & 43.7 & 17.1 & 10.9 & 38.0 & 25.6  \\
            UniVLA      & 1.8  & 46.2 & 69.9 & 69.0 & 81.0 & 21.2 & 31.9 & 45.9  \\
            $\pi_0$     & 13.8 & 6.0  & 58.8 & 85.0 & 81.4 & 79.0 & 68.9 & 56.1  \\
            $\pi_0$-FAST & 65.1 & 21.6 & 61.0 & 73.2 & 73.2 & 74.4 & 68.8 & 62.5  \\
            RIPT-VLA    & 55.2 & 31.2 & 77.6 & 88.4 & 91.6 & 73.5 & 74.2 & 70.2 \\
            OpenVLA-OFT & 56.4 & 31.9 & 79.5 & 88.7 & \textbf{93.3} & 75.8 & 74.2 & 71.4 \\
            AVA-VLA     & 55.5 & 25.9 & 85.6 & 95.5 & 88.9 & 78.0 & 74.1 & 71.9 \\
            MergeVLA~\cite{fu2025mergevla}   & 58.2 & 35.6 & 70.2 & 93.1 & 94.2 & 78.5 & 75.3 & 72.2 \\
            $\pi_{0.5}$       & 53.0 & \textbf{50.3} & 65.7   & 83.1   & 77.3   & 53.2   & 72.7  & 65.0      \\
            \rowcolor{oursvivid}
            MINT-30M & 61.4 & 41.2 & 61.6   & 92.2 & 77.1 & 76.5 & 76.2 & 69.5    \\
            \rowcolor{oursvivid}
            \textbf{MINT-4B}      & \textbf{72.2} & 42.4 & \textbf{85.8} & \textbf{96.6} & 88.9 & \textbf{90.1} & \textbf{84.6} & \textbf{80.1}  \\
            \midrule
            \multicolumn{9}{l}{\textit{Trained with LIBERO Plus}} \\
            OpenVLA-OFT+ & 92.8 & 30.3 & \textbf{85.8} & 94.9 & 93.9 & 89.3 & 77.6 & 80.7 \\
            $\pi_{0.5}$+ & 67.2 & 42.4 & 59.4 & 75.8 & 74.9 & 72.6 & 64.5 & 65.3  \\
            \rowcolor{oursvivid}
            \textbf{MINT-4B+}   & \textbf{95.6} & \textbf{44.6} & 84.7 & \textbf{95.1} & \textbf{94.5} & \textbf{95.2} & \textbf{78.7} & \textbf{84.1}  \\
            \bottomrule
        \end{tabular}
    }
\end{table*}

\begin{table*}[t]
    \centering
    \caption{Suite-wise generalization on LIBERO-PLUS under different perturbation factors}
    \label{tab:libero_plus_breakdown}
    \renewcommand{\arraystretch}{1.2}
    \setlength{\tabcolsep}{6pt}
    \resizebox{\linewidth}{!}{
    \begin{tabular}{ll ccccccc}
        \toprule
        \makecell[l]{Method}
        & \makecell[l]{Benchmark}
        & \makecell[c]{Camera\\Viewpoints} 
        & \makecell[c]{Robot\\Initial States} 
        & \makecell[c]{Language\\Instructions} 
        & \makecell[c]{Light\\Conditions} 
        & \makecell[c]{Background\\Textures} 
        & \makecell[c]{Sensor\\Noise} 
        & \makecell[c]{Objects\\Layout} \\
        \midrule
        
        \multirow{4}{*}{MINT-30M}
        & SPATIAL & 66.60 & 45.21 & 84.23 & 93.17 & 75.61 & 81.17 & 87.90 \\
        & OBJECT  & 57.98 & 35.29 & 74.21 & 99.83 & 73.39 & 68.45 & 77.72 \\
        & GOAL    & 74.36 & 48.48 & 45.77 & 96.60 & 94.49 & 86.10 & 64.22 \\
        & LONG      & 49.71 & 38.92 & 45.32 & 82.29 & 68.94 & 72.13 & 77.84 \\
        \midrule
        
        \multirow{4}{*}{MINT-4B}
        & SPATIAL & 77.99 & 50.14 & 91.92 & 96.92 & 97.67 & 92.60 & 97.40 \\
        & OBJECT  & 81.44 & 37.69 & 97.32 & 99.83 & 93.55 & 99.76 & 85.61 \\
        & GOAL    & 72.85 & 34.23 & 61.34 & 95.83 & 82.03 & 83.34 & 65.41 \\
        & LONG      & 58.67 & 48.60 & 95.17 & 96.94 & 83.56 & 85.73 & 93.59 \\
        \midrule
        
        \multirow{4}{*}{MINT-4B+}
        & SPATIAL & 96.54 & 50.86 & 90.51 & 98.46 & 97.67 & 97.29 & 96.75 \\
        & OBJECT  & 99.37 & 40.48 & 99.15 & 99.83 & 99.80 & 99.17 & 79.78 \\
        & GOAL    & 89.83 & 34.67 & 56.46 & 89.56 & 88.26 & 90.47 & 55.53 \\
        & LONG      & 96.40 & 53.59 & 95.69 & 94.16 & 93.08 & 96.10 & 86.70 \\
        
        \bottomrule
    \end{tabular}
    }
\end{table*}

\subsection{Statement on the Use of Large Language Models}
The manuscript was prepared with limited editorial assistance from large language models (LLMs). This assistance was restricted to improving the quality of the written expression, including grammar, sentence flow, and clarity. The underlying research concepts, methods, and conclusions were conceived, developed, and validated exclusively by the authors.

\clearpage

\end{document}